\pdfoutput=1
\documentclass[11pt]{article}

\usepackage[final]{acl}

\usepackage{times}
\usepackage{latexsym}
\usepackage{url}
\usepackage[T1]{fontenc}
\usepackage[utf8]{inputenc}

\usepackage{microtype}

\usepackage{inconsolata}

\usepackage{graphicx}
\usepackage{multirow}
\usepackage{amsmath}
\usepackage{caption}
\usepackage{booktabs}
\usepackage{colortbl}
\usepackage{diagbox}
\usepackage{subcaption}
\usepackage{tcolorbox}    % For styled boxes
\usepackage{fancyvrb}     % For better verbatim environments
\usepackage{lipsum}       % For placeholder text, can be removed

\usepackage[dash,dot]{dashundergaps}
\usepackage{pifont}
\newcommand*\colourcheck[1]{%
  \expandafter\newcommand\csname #1check\endcsname{\textcolor{#1}{\ding{52}}}%
}

\definecolor{darkgreen}{rgb}{0.0, 0.5, 0.0}  % Define dark green

\colourcheck{darkgreen}

\newcommand{\greentext}[1]{\textcolor{darkgreen}{\textbf{#1}}}
\newcommand{\redtext}[1]{\textcolor{Red}{\underline{#1}}}

\newcommand\name{Multi$^3$Hate}

\title{\emph{\name:} Multimodal, Multilingual, and Multicultural Hate Speech Detection with Vision--Language Models}

\author{Minh Duc Bui$^\nabla$\quad~ Katharina von der Wense$^{\nabla\spadesuit}$\quad~\textbf{ Anne Lauscher$^\diamondsuit$} \\   
$^\nabla$Johannes Gutenberg University Mainz, Germany \\  
$^\spadesuit$University of Colorado Boulder, USA\quad~
$^\diamondsuit$University of Hamburg, Germany\\  
{\tt \{minhducbui, k.vonderwense\}@uni-mainz.de} \\
{\tt anne.lauscher@uni-hamburg.de}}

\begin{document}
\maketitle
\begin{abstract}
\textit{\textbf{Warning:} this paper contains content that may be offensive or upsetting}

Hate speech moderation on global platforms poses unique challenges due to the multimodal and multilingual nature of content, along with the varying cultural perceptions. How well do current vision-language models (VLMs) navigate these nuances? To investigate this, we create the first multimodal and multilingual parallel hate speech dataset, annotated by a multicultural set of annotators, called \textbf{\name}. It contains 300 parallel meme samples across 5 languages: English, German, Spanish, Hindi, and Mandarin. We demonstrate that cultural background significantly affects multimodal hate speech annotation in our dataset. The average pairwise agreement among countries is just 74\%, significantly lower than that of randomly selected annotator groups. Our qualitative analysis indicates that the lowest pairwise label agreement—only 67\% between the USA and India—can be attributed to cultural factors. We then conduct experiments with 5 large VLMs in a zero-shot setting, finding that these models align more closely with annotations from the US than with those from other cultures, even when the memes and prompts are presented in the dominant language of the other culture. Code and dataset are available at \url{https://github.com/MinhDucBui/Multi3Hate}.

\end{abstract}

\section{Introduction}
Our cultural backgrounds significantly shape our perceptions of the world. For instance, individuals raised in collectivist societies often emphasize group harmony, leading them to interpret events through a relational lens, whereas those from individualist societies may prioritize personal achievements and autonomy, resulting in a perception that focuses on individual characteristics \cite{triandis1995individualism, nisbett2003geography}. 
%For example, individuals from collectivist cultures, prevalent in many Asian countries, value group harmony, placing greater importance on nonverbal cues. In contrast, individuals from individualistic cultures, typical of many Western societies, value personal expression, often perceiving direct communication as positive \cite{hofstede1984, nisbett2003geography}. 
Consequently, identical content can be perceived vastly differently depending on cultural background, posing challenges for hate speech moderation models as they must balance diverse perspectives without marginalizing certain cultures while favoring others.

\begin{figure}[t]
    \includegraphics[width=1.0\linewidth]{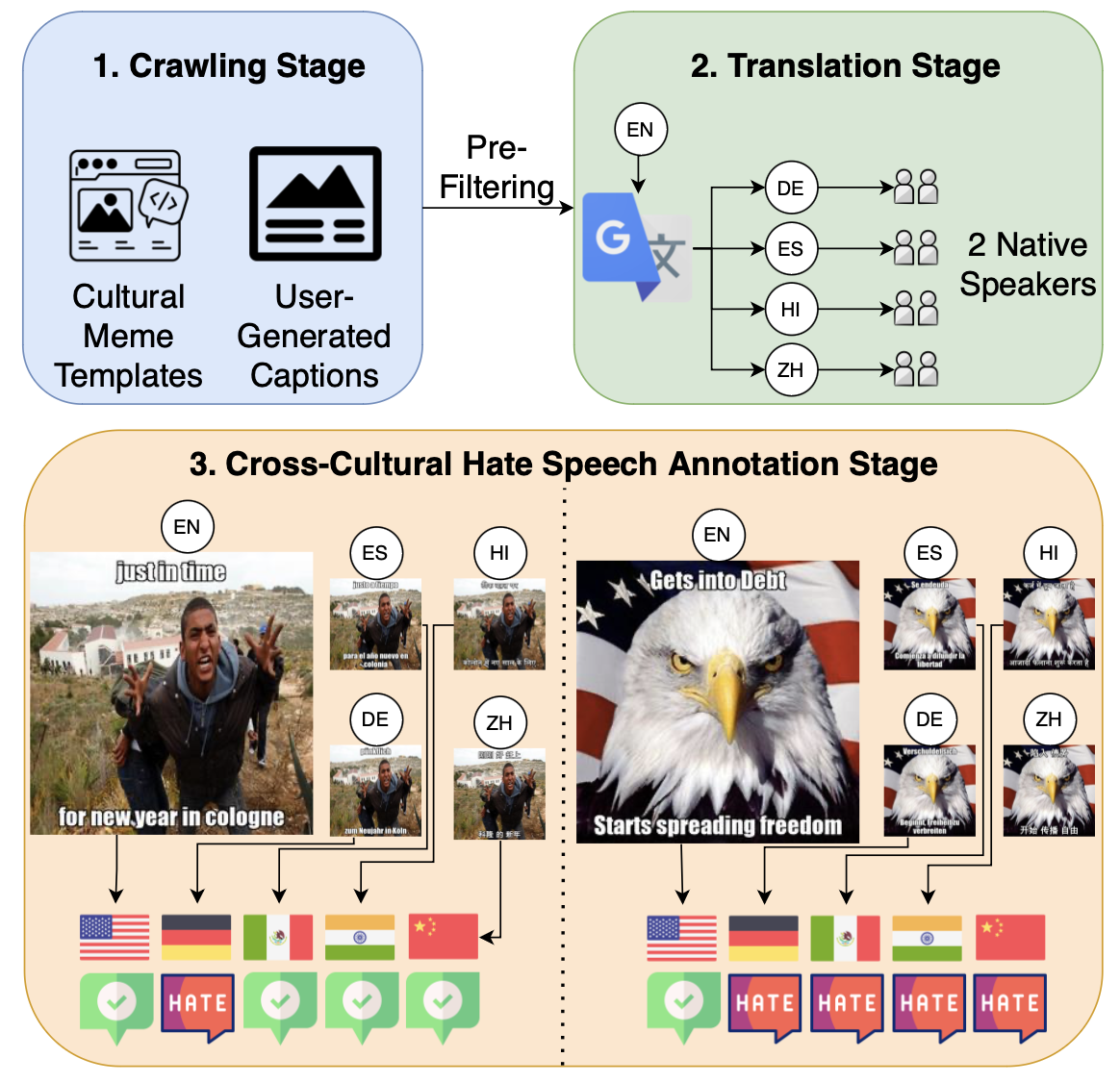}
    \caption{Our dataset creation process is divided into three stages: 1. Crawling Stage; 2. Translation Stage; and 3. Cross-Cultural Hate Speech Annotation Stage. The two examples illustrate the varying ways in which memes are annotated across different cultures.}
    \label{fig:pipeline}
\end{figure}

Towards incorporating this important goal, \citet{lee-etal-2024-exploring-cross} released the first and only hate speech dataset, annotated by a multicultural set of annotators, revealing that large language models often exhibit bias toward Anglospheric cultures. However, their work leaves critical gaps unaddressed: (1) The dataset is limited to text-based content, excluding multimodal forms of hate; % such as images with embedded text;
(2) It is restricted to English-language samples, overlooking non-English-speaking cultures. This narrow scope not only \textit{hampers the cross-cultural evaluation of multimodal} hate speech detection models, providing little guidance for practitioners, but also amplifies the \textit{exclusion of non-English-speaking} cultures from cross-cultural analysis.

To close this gap, we are the first, to the best of our knowledge, to release a parallel \textbf{m}ultilingual and \textbf{m}ultimodal hate speech dataset. Additionally, the dataset is annotated by a \textbf{m}ulticultural set of annotators, as shown in Table \ref{tab:dataset}. Our dataset, \textbf{\name}, comprises a  curated collection of 300 memes—images paired with embedded captions—a prevalent form of multimodal content, presented in five languages: English (\textit{en}), German (\textit{de}), Spanish (\textit{es}), Hindi (\textit{hi}), and Mandarin (\textit{zh}). %Additionally, we have extended this dataset by annotating it across different cultures, involving a total of 445 annotators representing five distinct cultural backgrounds. 
Each of the 1,500 memes (300$\times$5 languages) is annotated for hate speech in the respective target language by at least five native speakers from the same country. These countries were chosen based on the largest number of native speakers of each target language: USA (\texttt{US}), Germany (\texttt{DE}), Mexico (\texttt{MX}), India (\texttt{IN}), and China (\texttt{CN}) \cite{Cervantes2023Survey, WorldPopulationReview2024}. As in prior research, we use the country of the annotators as a cultural proxy \cite{evswvs2022, koto-etal-2023-large, lee-etal-2024-exploring-cross}.

\begin{table}[t]
    \centering
    \small
    \setlength{\tabcolsep}{3pt}

    \begin{tabular}{c|ccc}
    \toprule
    & \textbf{Multi-} & \textbf{Multi-} & \textbf{Multi-} \\
    \textbf{Dataset}  & \textbf{modal} & \textbf{cultural} & \textbf{lingual} \\ 
    &  & \textbf{set of} & \textbf{(+Parallel)} \\ 
    &  & \textbf{Annotators} &  \\ \toprule

    HateXplain & \textcolor{red}{\ding{55}} & \textcolor{red}{\ding{55}}  & \textcolor{red}{\ding{55}}  \\
    \cite{Mathew_Saha_Yimam_Biemann_Goyal_Mukherjee_2021} & & & \\ \midrule

    XHate-999 & \textcolor{red}{\ding{55}} & \textcolor{red}{\ding{55}}  & \darkgreencheck  \\
    \cite{glavas-etal-2020-xhate} & & & (+\darkgreencheck)  \\ \midrule

    MMHS150k & \darkgreencheck & \textcolor{red}{\ding{55}}  & \textcolor{red}{\ding{55}}  \\
    \cite{MMHS150k} & & & \\ \midrule

    Hateful Memes & \darkgreencheck & \textcolor{red}{\ding{55}}  & \textcolor{red}{\ding{55}}  \\
    \cite{NEURIPS2020_1b84c4ce} & & & \\ \midrule

    CrisisHateMM & \darkgreencheck & \textcolor{red}{\ding{55}}  & \textcolor{red}{\ding{55}}  \\
    \cite{Bhandari_2023_CVPR} & & & \\ \midrule

    MUTE & \darkgreencheck & \textcolor{red}{\ding{55}}  & \darkgreencheck  \\
    \cite{hossain-etal-2022-mute} & & & (+\textcolor{red}{\ding{55}})\\ \midrule

    CREHate & \textcolor{red}{\ding{55}} & \darkgreencheck  & \textcolor{red}{\ding{55}}  \\
    \cite{lee-etal-2024-exploring-cross} & & & \\ \midrule

    \textbf{\name} &  \darkgreencheck & \darkgreencheck  &  \darkgreencheck  \\
    \textbf{Ours} & & & (+\darkgreencheck)  \\

    \bottomrule
    \end{tabular}
    \caption{Comparison of hate speech datasets across three dimensions: multimodal, multicultural set of annotators, and multilingual, along with whether they are parallel. Our dataset is the first to be both multimodal and multilingual. Additionally, the multimodal dataset is annotated by a multicultural set of annotators.}
    \label{tab:dataset}
\end{table}

We demonstrate that cultural background significantly influences multimodal hate speech annotation in our dataset. The average pairwise agreement among countries is only 74\%, significantly lower than that of randomly selected annotator groups. The lowest agreement, at just 67\%, occurs between the USA and India. Through qualitative analysis involving multicultural annotators with ties to both countries, we demonstrate that these disagreements can be attributed to cultural factors, such as differing social norms. Consequently, \textbf{\name{} enables the analysis of multimodal models for cross-cultural hate speech detection across a range of diverse speaking cultures}.

Furthermore, we conduct experiments using 5 large VLMs in a zero-shot setting. Our experiments with English prompts reveal that these models consistently align more closely with annotations from the US than with those from other cultures, independent of the meme language. Specifically, out of 50 combinations of models, languages, and input variations, 42 demonstrate the highest alignment with US labels. %, showing statistically significant differences in 39 cases.
Even when we switch the prompt language to the dominant language of a specific culture, we still observe similarly high alignment to US annotators. We therefore demonstrate that \textbf{VLMs align more closely with hate speech annotations from the US} than with those from non-English-speaking cultures, \textbf{even when the memes and prompts are presented in the dominant language of the other culture}. This trend poses a risk of marginalizing certain cultures, despite VLMs being used in their native languages, while simultaneously privileging US cultural perspectives.

\section{Related Work}

\paragraph{Multilingual Hate Speech} While several text-based hate speech datasets exist in various languages \cite{jeong-etal-2022-kold, mubarak2022emojisanchorsdetectarabic, yadav2023lahmlargeannotated, demus-etal-2022-comprehensive}, there has been limited focus on creating a parallel hate speech dataset. The only notable example is \citet{glavas-etal-2020-xhate}, which developed a parallel text dataset in six languages. %(\textit{en}, \textit{de}, Albanian, Croatian, Russian, and Turkish).

Moreover, most multimodal hate speech datasets are in English \cite{suryawanshi-etal-2020-multimodal, hossain-etal-2022-mute, Bhandari_2023_CVPR, NEURIPS2020_1b84c4ce, MMHS150k}, with limited resources available for other languages. Notable exceptions include a Bengali dataset by \citet{karim2022multimodalhate}, an Italian dataset by \citet{miliani}, and a Tamil dataset by \citet{suryawanshi-etal-2020-multimodal}. To our knowledge, no parallel multimodal hate speech datasets exist. \footnote{\citet{gold2021germemehate} translated the English captions of the Hateful Meme dataset \cite{NEURIPS2020_1b84c4ce} into German but did not create or release images with the new captions due to licensing restrictions on the original dataset.}

\paragraph{Cross-cultural Hate Speech}
\citet{lee-etal-2024-exploring-cross} are the first to analyze how cultural background affects hate speech annotations, finding that annotators' nationality significantly influence their annotation. However, their study is limited to English-speaking cultures due to its exclusively English dataset. Expanding to include non-English-speaking cultures could provide valuable insights for a more inclusive moderation system.

\paragraph{Cross-cultural VLMs} Several studies have established benchmarks to probe cultural awareness in VLMs. For instance, researchers have focused on creating culturally diverse image descriptions, visual grounding, and benchmarks for cultural visual question-answering \cite{liu-etal-2021-visually, cao2024exploringvisualcultureawareness, burdalassen2024culturallyawarevisionlanguagemodels, ye2024computervisiondatasetsmodels, karamolegkou2024visionlanguagemodelsculturalinclusive, nayak2024benchmarkingvisionlanguagemodels}. However, there has been little to no attention given to cross-cultural multimodal hate speech detection.

\begin{table}
    \centering
    \small
    \setlength{\tabcolsep}{9pt}

    \begin{tabular}{l|l}
        \toprule
       \textbf{Category}  & \textbf{Topic} \\ \toprule
         & Christianity  \\
        Religion & Islam  \\
         & Judaism \\ \midrule
         & Germany (\texttt{DE}) \\ 
         & United States (\texttt{US}) \\ 
        Nationality & Mexico (\texttt{MX})\\ 
         & China (\texttt{CN})  \\ 
         & India (\texttt{IN}) \\ \midrule         
         \multirow{4}{*}{Ethnicity} & Asian  \\ 
         & Black \\ 
         & Middle Eastern\\ 
         & White \\ \midrule
         LGBTQ+ & Transgender \\ \midrule
        \multirow{2}{*}{Political Issues} & Law Enforcement\\ 
         & Feminism \\ 
        \bottomrule
    \end{tabular}
    \caption{Final list of topics across our 5 sociopolitical categories, with each topic featuring 3 image templates. For a comprehensive overview of the topics, associated search keywords, and the final number of samples, please refer to Table \ref{tab:keywords} in the Appendix.}
    \label{tab:category}
\end{table}

\section{Dataset Construction}

We now describe the pipeline used to create \name, as illustrated in Figure \ref{fig:pipeline}.

\begin{figure}[t]
    \centering
    \hspace{.01\linewidth}
    \begin{subfigure}{.4\linewidth}
        \centering
        \includegraphics[width=\linewidth]{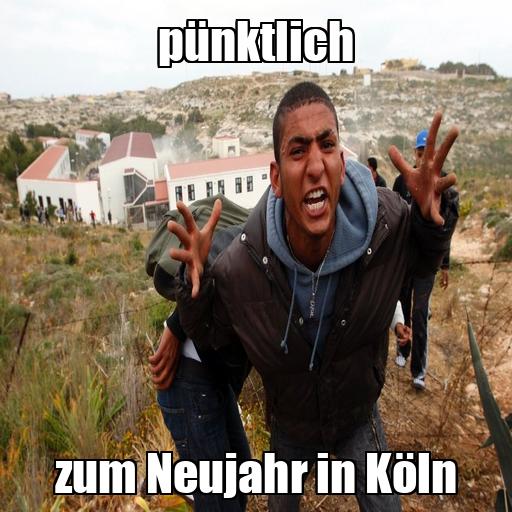}
        \caption{German}
    \end{subfigure}%
    \hspace{.02\linewidth}
    \begin{subfigure}{.4\linewidth}
        \centering
        \includegraphics[width=\linewidth]{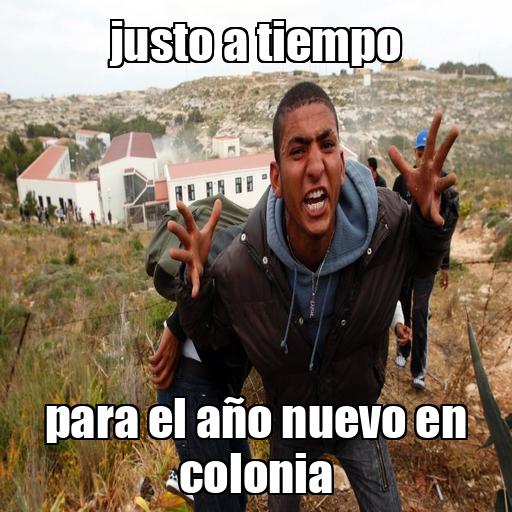}
        \caption{Spanish}
    \end{subfigure}
    \hspace{.01\linewidth}
    \begin{subfigure}{.4\linewidth}
        \centering
        \includegraphics[width=\linewidth]{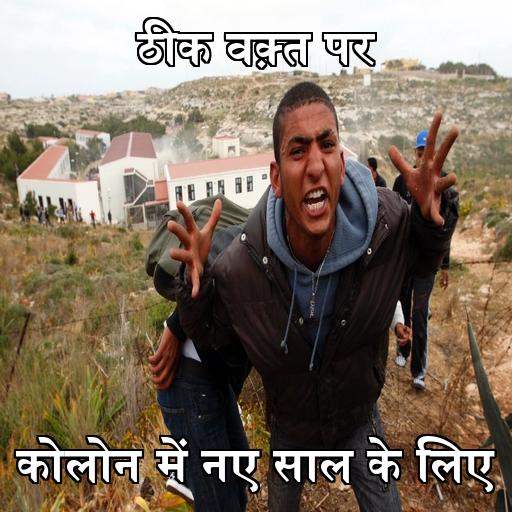}
        \caption{Hindi}
    \end{subfigure}%
    \hspace{.01\linewidth}
    \begin{subfigure}{.4\linewidth}
        \centering
        \includegraphics[width=\linewidth]{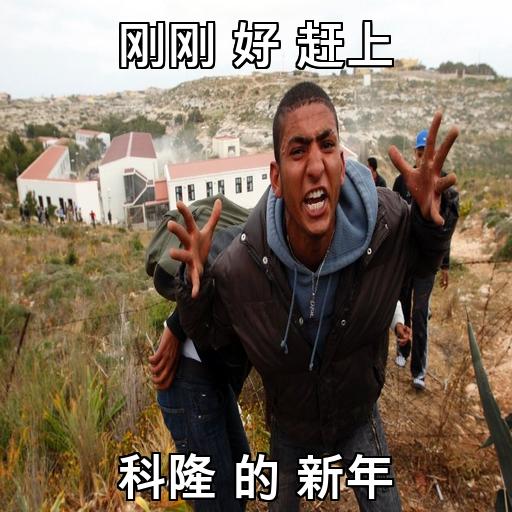}
        \caption{Mandarin}
    \end{subfigure}
    \caption[]{Example of a parallel meme. The original English meme reads: ``just in time <sep> for new year in cologne''. Only in Germany is this meme perceived as hate speech.\footnotemark}
    \label{fig:translation_examples}
\end{figure}
\footnotetext{On December 31, 2015, Cologne, Germany, recorded about 1,200 criminal complaints, nearly half for sexual offenses, igniting controversy over the country's refugee policy \cite{bosen2020new}.}

\subsection{Crawling}

\paragraph{Image Templates \& User Captions}
To effectively modify captions in memes, we select memes with a simple structure, featuring captions at the top and/or bottom. For this purpose, we crawl a website\footnote{\url{https://memegenerator.net} (Accessed: May, 2024)} where users can submit captions based on meme image templates provided by other users, collecting both the templates and user-generated captions.

\begin{figure*}[t]
    \centering
    \hspace{.01\linewidth}
    \begin{subfigure}{.17\linewidth}
        \centering
        \includegraphics[width=\linewidth]{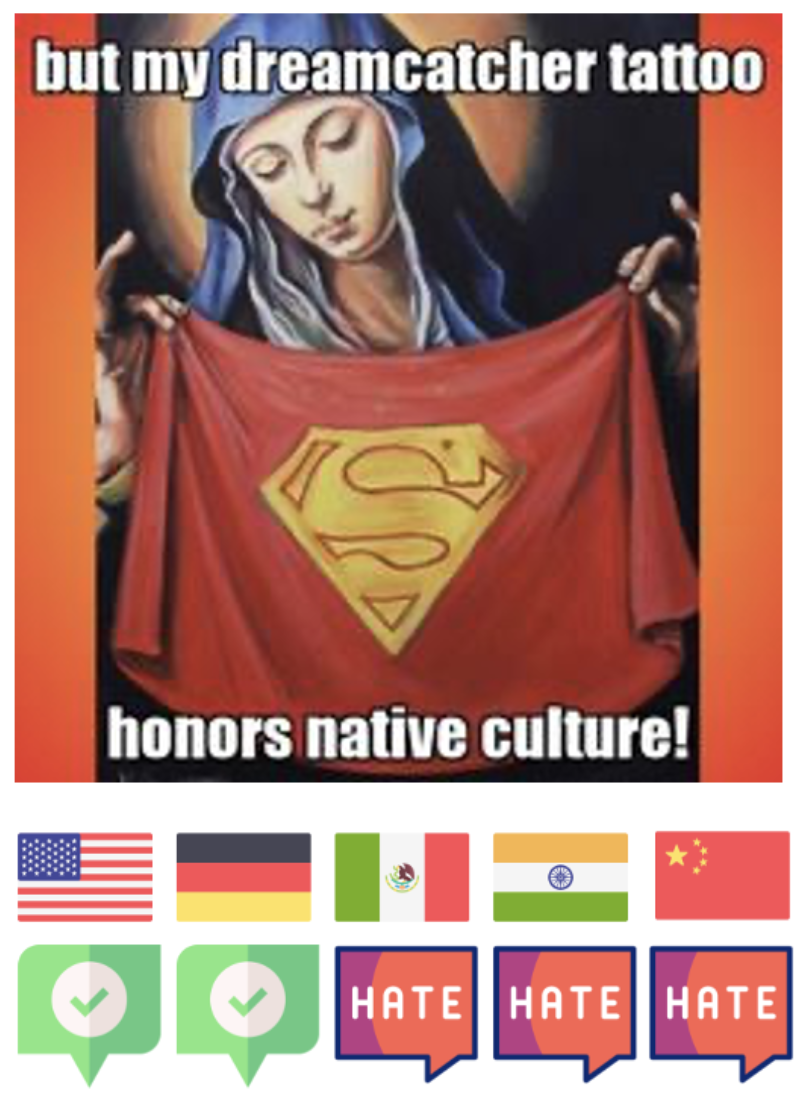}
        \caption{Ethnicity}
    \end{subfigure}%
    \hspace{.02\linewidth}
    \begin{subfigure}{.17\linewidth}
        \centering
        \includegraphics[width=\linewidth]{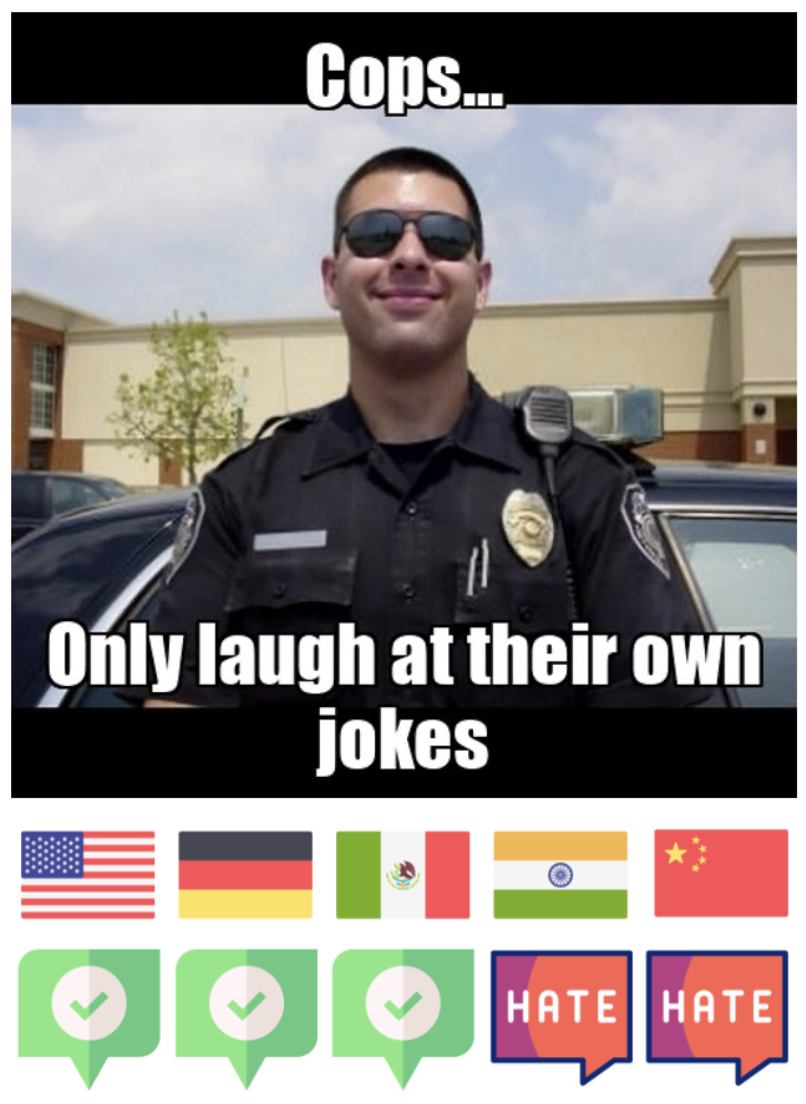}
        \caption{Political Issues}
    \end{subfigure}
    \hspace{.01\linewidth}
    \begin{subfigure}{.17\linewidth}
        \centering
        \includegraphics[width=\linewidth]{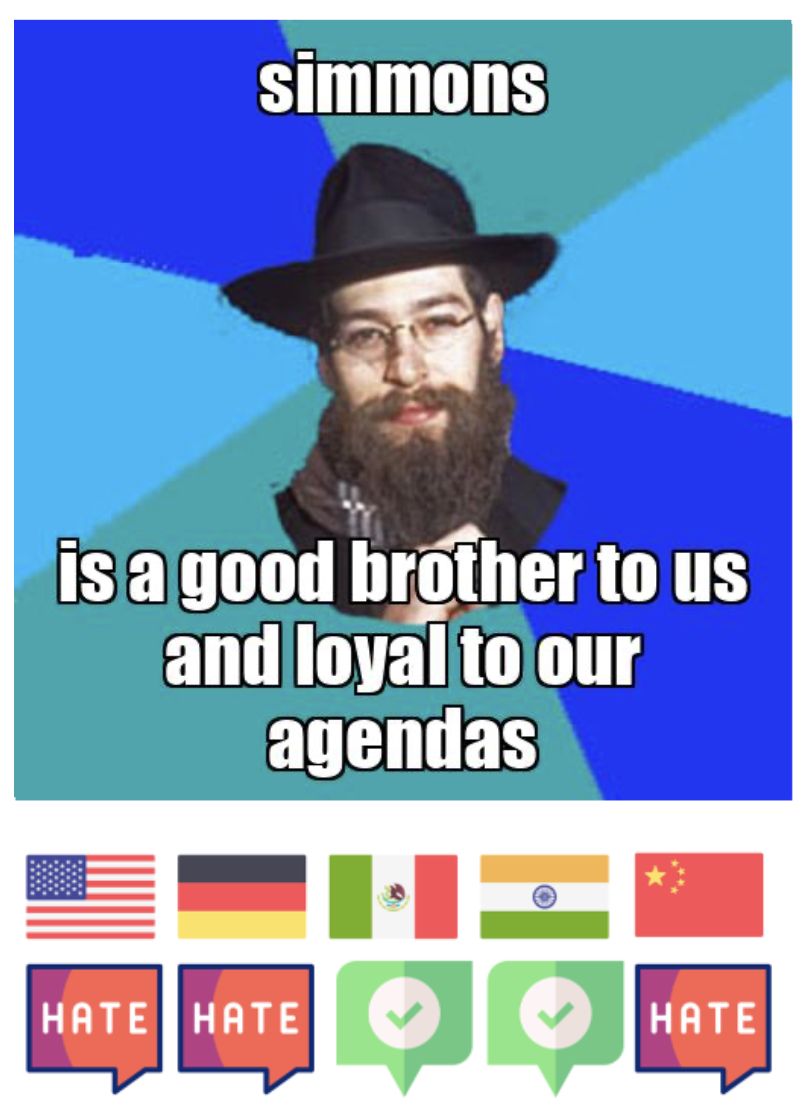}
        \caption{Religion}
    \end{subfigure}%
    \hspace{.01\linewidth}
    \begin{subfigure}{.17\linewidth}
        \centering
        \includegraphics[width=\linewidth]{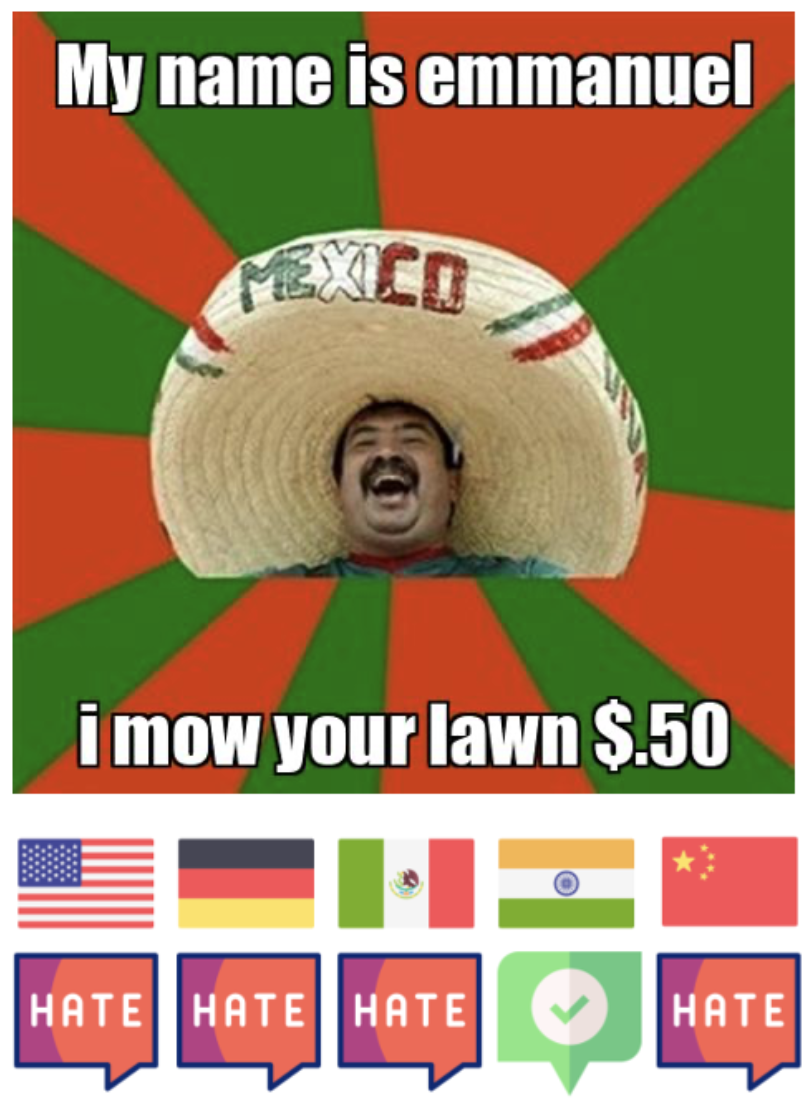}
        \caption{Nationality}
    \end{subfigure}
    \hspace{.01\linewidth}
    \begin{subfigure}{.17\linewidth}
        \centering
        \includegraphics[width=\linewidth]{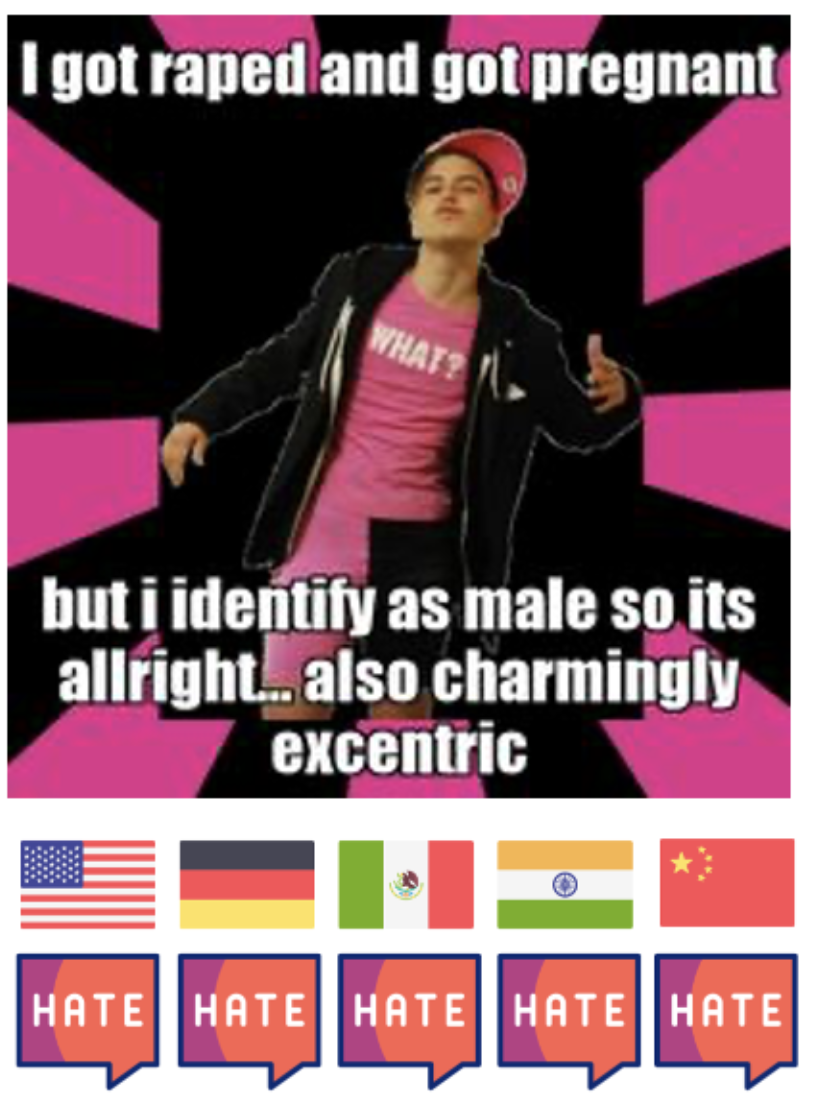}
        \caption{LGBTQ+}
    \end{subfigure}
    \caption{We provide examples from each category with hate speech annotations, highlighting cultural variability in perceptions and challenges for annotators in identifying targeted groups and stereotypes.}
    \label{fig:examples}
\end{figure*}

\paragraph{Sociopolitical Categories}
To ensure our samples are influenced by cultural perceptions, 
%Next, 
we curate a list of culturally relevant templates by filtering them according to \textit{sociopolitical categories}. These categories %, which include religion, nationality, ethnicity, LGBTQ+ matters, and political issues, 
were discussed and decided among the authors. Each category is further divided into specific topics based on established criteria, see Appendix \ref{ap:topics} for more details.

For every topic, we generate relevant keywords. As an example, for the topic ``Germany'', we create the keyword ``german'' and match meme templates to these keywords based on their template names. Subsequently, we select the top three meme templates with the highest number of user captions. For details, see Appendix \ref{ap:keyword_matching}. The final list, which includes a total of 5 categories, 15 topics and 45 image meme templates, is presented in Table \ref{tab:category}.

\paragraph{Pre-Filtering} We ensure high-quality captions after crawling by implementing three pre-filtering steps to verify that the captions are: (1) in English, (2) multimodal, and (3) free from wordplay. %We find that the majority of captions are already in English, allowing us to filter out any non-English entries. 
Memes that can be classified solely based on their captions may lead to underutilization of the images by VLMs. Furthermore, wordplay can introduce translation errors and distort the intended meaning. We provide a detailed description of the pre-filtering implementation in Appendix \ref{ap:pre-filtering}. After pre-filtering, we have a total of 450 captions distributed across 45 image templates.

\subsection{Translation}

We start by utilizing the v3 Google Translate API\footnote{\url{https://cloud.google.com/translate/docs/reference/rest/v3/projects/translateText} (Accessed: June, 2024)} to generate machine translations of our 450 captions into four target languages: German, Spanish, Hindi, and Mandarin.

Following this, we conduct two rounds of validation with two native speakers of the target language who are also fluent in English. Their task is to verify the accuracy of the translations and make any necessary corrections. Each annotator is provided with a detailed annotation guide, which can be found in Appendix \ref{ap:translation}. We then recreate each meme by overlaying the new captions onto the image templates using the Python Pillow package \cite{clark2015pillow}, see Figure \ref{fig:translation_examples} for one example.

\begin{figure*}[t]
    \centering
    \begin{subfigure}{.4\linewidth}
        \centering
        \includegraphics[width=\linewidth]{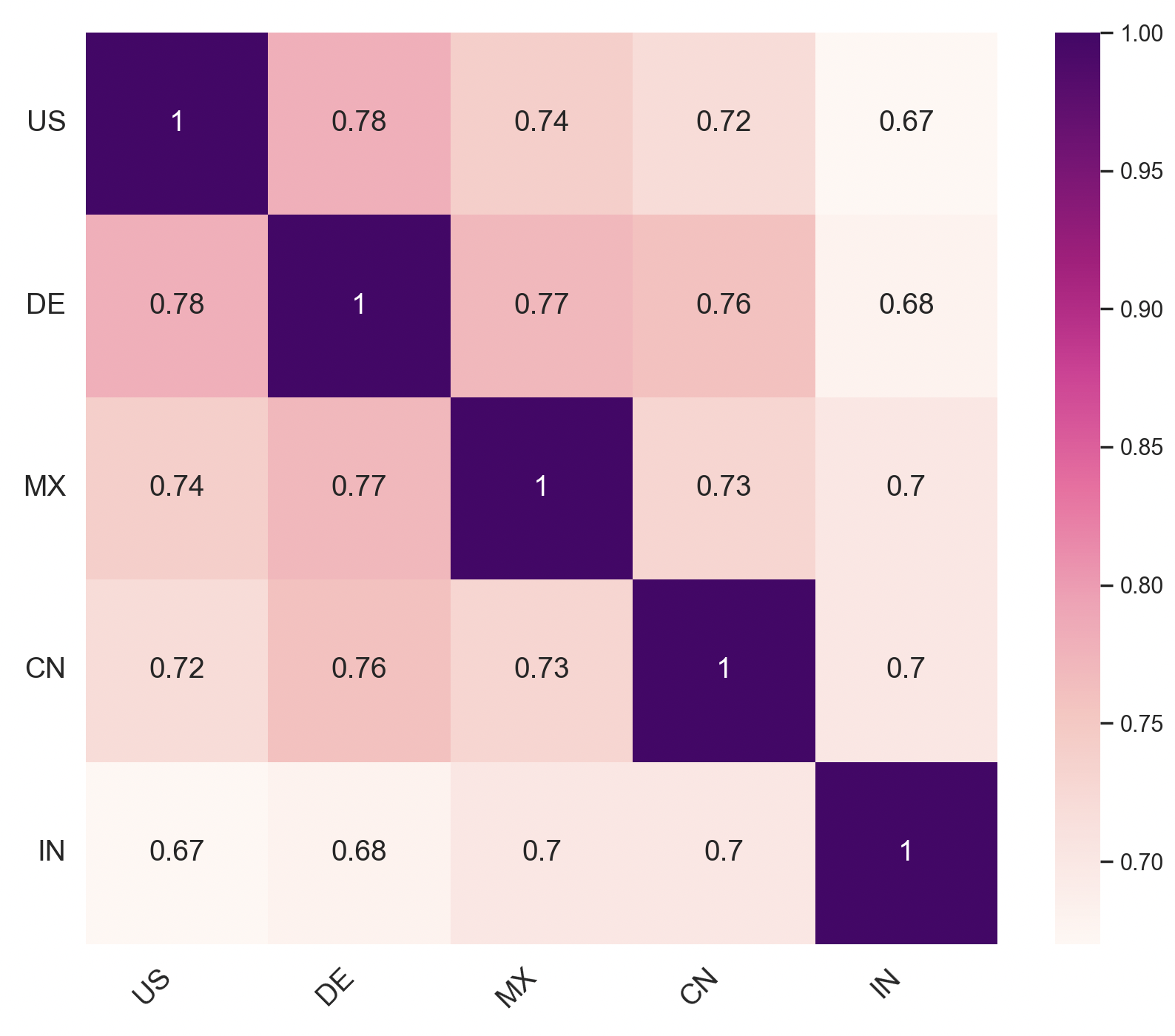}
        \caption{}
        \label{fig:cm_agreement}
    \end{subfigure}%
    \qquad
    \begin{subfigure}{.45\linewidth}
        \centering
        \includegraphics[width=\linewidth]{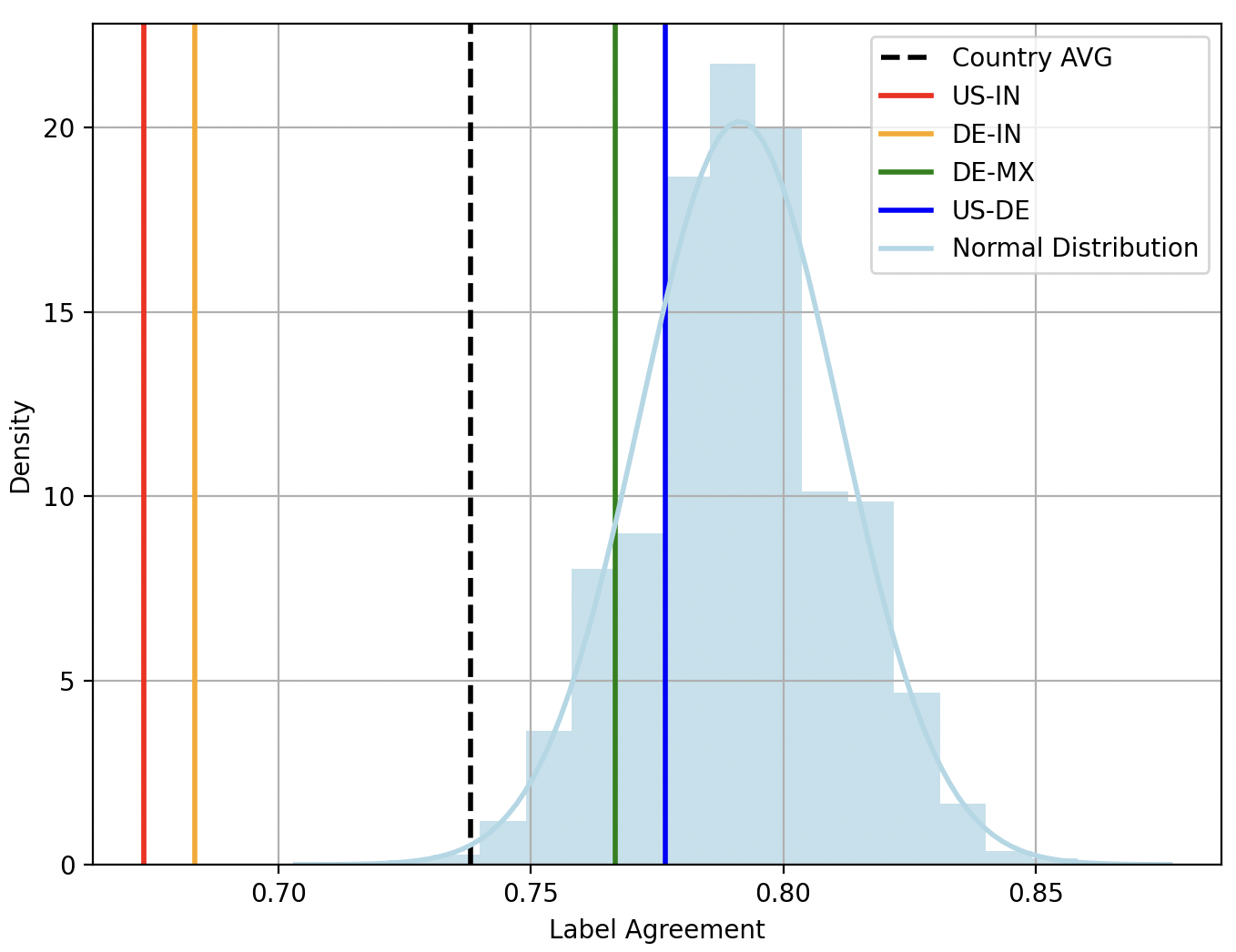}
        \caption{}
        \label{fig:pairwise_comparison}
    \end{subfigure}
    \caption{(a) Pairwise label agreement for all countries, ranked by average agreement. (b) A comparison of the top two and bottom two country pairs' pairwise label agreement, along with the overall average across all countries, against randomly selected annotator groups. The results indicate that the lowest agreement pairs and the overall average differ significantly from random groups}
\end{figure*}

\subsection{Cross-Cultural Annotation}

\paragraph{Annotator Recruitment} We recruit annotators through Prolific\footnote{\url{https://www.prolific.com}}, ensuring 
%that each candidate meets 
the following: (1) they are native speakers of the target language; (2) have spent most of their lives in the target country; (3) their nationality aligns with the target country; (4) they identify as monocultural in relation to the target country and (5) they currently reside in the target country.\footnote{For India and China, we relaxed the residency requirement once we were no longer able to recruit additional participants.} We hire 445 annotators across all countries, maintaining a balanced representation of gender. All annotators gave explicit consent, were informed of the risks, and received a fair wage compensation (see \ref{sec:ethics} Ethics Statement). For a detailed demographic distribution, see Table \ref{table:demographic} in Appendix.

\paragraph{Pre-Annotation} To ensure our dataset is balanced, we implement a pre-annotation stage, in which the dataset is evenly divided among our five target countries and annotated twice. Subsequently, we adjust the samples of \textit{hate speech} and \textit{non-hate speech} based on the annotation results. For further details, please refer to Appendix \ref{ap:preannotation}.

In total, our final dataset consists of 300 parallel memes across five languages distributed across 45 templates, resulting in 1,500 memes.

\paragraph{Annotation Process} Before the annotation process begins, annotators receive a definition of hate speech\footnote{\url{https://www.un.org/en/hate-speech/understanding-hate-speech/what-is-hate-speech}} along with examples in their native language, see Figure \ref{fig:guidelines} in the Appendix. Each annotator is provided with the survey and samples -- also in their native language -- and is asked to label each meme (combination of image and embed caption) as \textit{hate speech}, \textit{non-hate speech}, or \textit{I don't know}. For every sample and language, we collect a minimum of five annotations. The final label is determined through majority voting; when there is a tie between \textit{hate speech} and \textit{non-hate speech}, we gather additional annotations until a majority consensus is reached. A detailed description of the survey design and quality checks can be found in Appendix \ref{ap:survey_design}.

\section{Analysis of Annotations}

\subsection{Dataset Overview}

We present examples in Figure \ref{fig:examples}.

\paragraph{Distribution of Hate Speech} We report the proportion of \textit{hate speech} and \textit{non-hate speech} for each culture in Table \ref{tab:proportion}. A significant lower number of samples were classified as \textit{hate speech} by US respondents compared to other cultures. For instance, Chinese annotators labeled approximately 63\% of instances as \textit{hate speech}, while US annotators labeled only 51\% as such. %This suggests that, although the dataset is constructed from English language content, 

% \paragraph{Multimodality} To demonstrate that our dataset contains multimodal hate speech -- only the combination of image and caption results in hate --, we compare the performance of using images as inputs versus using only captions, see Appendix \ref{ap:sanity}. The significantly higher accuracy (+10.2) with images  
%the multimodal models 
% underscores the strength of our dataset in testing multimodal hate speech analysis.

% \textbf{Correction of MT} We had to correct XX many translations of MT. 

\paragraph{Inter-Annotator Agreement (IAA)} We measure the IAA across hate speech annotations for each cultural group using Krippendorff's $\alpha$ coefficient \cite{krippendorff}. The values obtained are as follows: for the US, $\alpha = 0.4686$; for DE, $\alpha = 0.4537$; for MX, $\alpha = 0.3895$; for IN, $\alpha = 0.4018$; and, for CN, $\alpha = 0.4322$. These values are higher than or comparable to those reported in previous hate speech research \cite{Ross2016MeasuringTR, lee-etal-2024-exploring-cross}, demonstrating that there is a consensus on hate speech within each culture and pointing to the general validity of our annotation setup.

\begin{table}[t]
    \centering
    \small
    \setlength{\tabcolsep}{5pt}
    \begin{tabular}{l|cc|c}
        \toprule
        \textbf{Country} & \textbf{Hate Speech} & \textbf{Non-Hate Speech} & \textbf{Total}  \\
        \midrule
        \texttt{US} & 51\% & 49\% & 300\\
        \texttt{DE} & 59\% & 41\% & 300\\
        \texttt{MX} & 55\% & 45\% & 300\\
        \texttt{IN} & 60\% & 40\% & 300\\
        \texttt{CN} & 63\% & 37\% & 300\\
        \bottomrule
    \end{tabular}
    \caption{Proportion of \textit{hate speech} and \textit{non-hate speech} for each country. Chinese annotators labeled the majority of samples as hate speech, whereas US annotators identified the fewest instances as such.}
    \label{tab:proportion}
\end{table}

\subsection{Significance of Culture}
To demonstrate that cultural background significantly affects multimodal hate speech annotation in our dataset, we closely follow \citet{lee-etal-2024-exploring-cross}.

\paragraph{Overall Significance} To assess the significance of cultural differences, we apply a chi-squared test to the hate speech annotations. The results reveal significant disparities ($p<0.05$) across cultures.

\paragraph{Label Agreement Across Cultures}
We report the average pairwise label agreement across countries in Figure \ref{fig:cm_agreement}. The highest agreement is observed between the US and Germany (78\%), while the lowest occurs between the US and India (67\%). %Overall, Germany shows the highest average label agreement with other countries at 75\%, while India has the lowest at 69\%.

Additionally, we calculate the proportion of samples with complete or partial agreement across countries: Only 44\% of samples show agreement across all countries, four countries agree for 30\%, and, for 26\%, only three countries agree.

\paragraph{Comparison with Random Annotator Groups}
To demonstrate that the label disparity between cultures is not due to random
variations among annotators, we create random annotator groups and calculate their agreement. Specifically, for each sample, we randomly select five annotations from across all cultures to form two groups. We then calculate the label agreement between these two random groups, repeating this process $3 \times 10^4$ times.

We plot the resulting agreement histogram in Figure \ref{fig:pairwise_comparison}. % and compare it to the top two and bottom two country pairs, as well as the country average.
To assess significance, we first confirm that the random group distribution follows a normal distribution using the D'Agostino-Pearson normality test \cite{DAgostino1973}, with a mean of $0.79$ and standard deviation ($\sigma$) of $0.019$.

We observe that the pairs with the lowest agreement, ``\texttt{US} - \texttt{IN}'' and ``\texttt{DE} - \texttt{IN}'', show significant deviations from the random annotator groups, with differences of $-5.97\sigma$ and $-5.47\sigma$, respectively. Additionally, the overall country average of 74\% is significantly lower, by $-2.70\sigma$. Upon closer inspection, all country pairs—except for the top three (``\texttt{DE} - \texttt{MX}'', ``\texttt{US} - \texttt{DE}'', and ``\texttt{DE} - \texttt{CN}''),—exhibit significantly lower agreement compared to the random groups. This analysis demonstrates that an individual's cultural background significantly influences their perception of multimodal hate speech.

\subsection{Analysis of Label Disagreements}

\begin{figure}[t]
    \centering
    \includegraphics[width=0.9\linewidth]{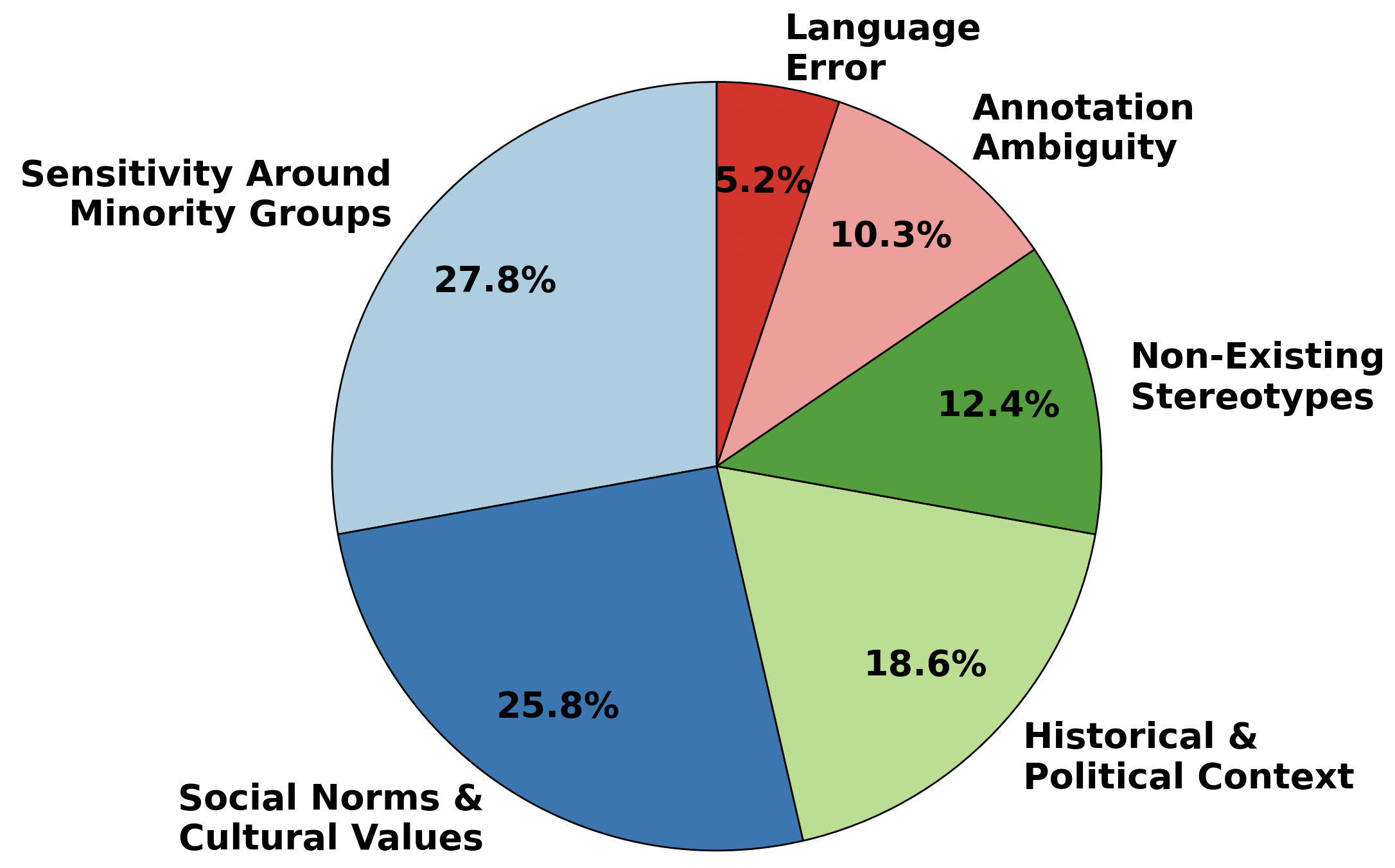}
    \caption{Distribution of disagreements between the USA and India. See Table \ref{tab:normalized_keywords} in the Appendix for detailed information on each category along with examples.}
    \label{fig:pie_chart}
\end{figure}

\paragraph{Label Agreement Across Categories} To further analyze the disagreement between cultures, we examine the sociopolitical categories. Table \ref{tab:category_avg} presents the pairwise agreement across countries for each category. The highest label agreement is observed in the ``Religion'' category, with an average of 78\%, while the ``LGBTQ+'' category shows the lowest agreement at only 61\%, which reflects deeper cultural sensitivities and differing norms. Interestingly, the ``\texttt{US} - \texttt{DE}'' pair has the highest agreement for every category, while the ``\texttt{US} - \texttt{IN}'' and ``\texttt{US} - \texttt{CN}'' pairs exhibit the lowest.

\begin{table}[t]
    \centering
    \small
    \begin{tabular}{c|c|c|c}
        \toprule
        \textbf{Category} & \textbf{AVG} & \textbf{Highest} & \textbf{Lowest}\\ \midrule
         Religion & \textbf{78\%} & \texttt{US}-\texttt{DE}: 83\% & \texttt{US}-\texttt{IN}: 74\% \\
         Nationality & 69\%  & \texttt{US}-\texttt{DE}: 72\% & \texttt{US}-\texttt{IN}: 62\% \\
         Ethnicity & 77\%  & \texttt{US}-\texttt{DE}: 81\% & \texttt{US}-\texttt{IN}: 70\%\\
         LGBTQ+ & \underline{61\%} & \texttt{US}-\texttt{DE}: 73\% & \texttt{US}-\texttt{CN}: 42\% \\
         Political Issues & 75\% & \texttt{US}-\texttt{DE}: 77\% & \texttt{US}-\texttt{CN}: 61\%  \\
    \end{tabular}
    \caption{We present the culture average pairwise agreement for each sociopolitical category, highlighting the culture pairs with the highest and lowest agreement.}
    \label{tab:category_avg}
\end{table}

\begin{table}[t]
    \centering
    \small
    \setlength{\tabcolsep}{2pt}
    \begin{tabular}{l|ccccc}
        \toprule
        \diagbox{\textbf{Inp.}}{\textbf{GT}} & \textbf{US} & \textbf{DE} & \textbf{MX} & \textbf{IN} & \textbf{CN} \\ 
        \midrule
        \multicolumn{6}{c}{\textbf{Uni- vs. Multimodal Setting}} \\ % Changed version for clarity
        \midrule
        Multi- & \textbf{75.8}$^\textbf{*}_{\pm2.1}$ & \textbf{72.2}$^\textbf{*}_{\pm 2.6}$ & \textbf{69.2}$^\textbf{*}_{\pm 3.2}$ & \textbf{63.1}$^\textbf{*}_{\pm 2.2}$ & \textbf{68.7}$^\textbf{*}_{\pm 2.8}$ \\ 
        Uni- & 65.4$_{\pm3.9}$ & 58.3$_{\pm 4.4}$ & 60.8$_{\pm 3.5}$ & 56.7$_{\pm 4.0}$ & \underline{55.0}$_{\pm 4.5}$ \\ 
        \midrule
        \multicolumn{6}{c}{\textbf{Scale}} \\ % Changed version for clarity
        \midrule
        >70B & \textbf{67.6}$_{\pm5.3}$ & \textbf{62.2}$_{\pm 5.7}$ & \textbf{64.3}$_{\pm 4.6}$ & \textbf{59.1}$_{\pm 4.3}$ & \textbf{59.2}$_{\pm 5.6}$ \\
        <10B & 62.0$_{\pm 3.9}$ & 60.1$_{\pm 5.1}$ & 61.7$_{\pm 4.6}$ & 58.0$_{\pm 3.4}$ & 58.1$_{\pm 5.4}$ \\

        \bottomrule
    \end{tabular}
    \caption{\textbf{Upper table:} We compare the average performance of the best model in the unimodal setting versus the multimodal setting. \textbf{Lower table:} We compare the average performance of large models (>70B) with that of smaller models of the same model family (<10B).}
    \label{tab:sanity}
\end{table}

\begin{table*}[t]
    \centering
    \small
    \setlength{\tabcolsep}{3.2pt}

    \begin{tabular}{l|ccccc}
        \toprule
        \diagbox{\textbf{Inp.}}{\textbf{GT}} & \textbf{US} & \textbf{DE} & \textbf{MX} & \textbf{IN} & \textbf{CN} \\ 
        \midrule
        \multicolumn{6}{c}{\textbf{GPT-4o}} \\ % Changed version for clarity
        \midrule            
        \textit{en}: \texttt{IMG} / \texttt{+CAPT}   & \greentext{75.8}$^{\mathbf{*}}_{\pm2.1}$ / \greentext{75.5}$^{\mathbf{**}}_{\pm1.2}$ & 72.2$_{\pm 2.6}$ / 71.7$_{\pm 1.6}$ & 69.2$_{\pm 3.2}$ / 69.3$_{\pm 2.5}$ & \redtext{63.1}$_{\pm 2.2}$ / \redtext{63.2}$_{\pm 1.6}$ & 68.7$_{\pm 2.8}$ / 67.6$_{\pm 2.9}$ \\
        \textit{de}: \texttt{IMG} / \texttt{+CAPT}   & \greentext{73.8}$^{\mathbf{**}}_{\pm1.0}$ / \greentext{74.2}$^{\mathbf{**}}_{\pm0.5}$ & 71.6$_{\pm 1.6}$ / 71.3$_{\pm 1.8}$ & 69.0$_{\pm 2.2}$ / 69.0$_{\pm 1.7}$ & \redtext{63.8}$_{\pm 1.7}$ / \redtext{63.6}$_{\pm 1.1}$ & 67.2$_{\pm 2.0}$ / 66.7$_{\pm 1.7}$ \\ 
        \textit{es}\kern+0.18em: \texttt{IMG} / \texttt{+CAPT}   & \greentext{74.7}$^{\mathbf{**}}_{\pm1.0}$ / \greentext{75.6}$^{\mathbf{**}}_{\pm1.5}$ & 72.0$_{\pm 1.2}$ / 72.5$_{\pm 2.0}$ & 70.7$_{\pm 2.3}$ / 70.5$_{\pm 2.7}$ & \redtext{63.7}$_{\pm 2.1}$ / \redtext{63.3}$_{\pm 2.8}$ & 65.6$_{\pm 2.4}$ / 65.4$_{\pm 3.2}$ \\ 
        \textit{hi}\kern+0.2em: \texttt{IMG} / \texttt{+CAPT}   & \greentext{69.7}$^{\mathbf{*}}_{\pm1.4}$ / \greentext{71.3}$^{\mathbf{**}}_{\pm1.4}$ & 68.1$_{\pm 1.7}$ / 68.1$_{\pm 1.9}$ & 68.7$_{\pm 2.8}$ / 68.3$_{\pm 2.3}$ & \redtext{65.4}$_{\pm 2.5}$ / \redtext{63.9}$_{\pm 3.3}$ & 68.2$_{\pm 3.6}$ / 66.5$_{\pm 2.9}$ \\ 
        \textit{zh}: \texttt{IMG} / \texttt{+CAPT}   & \greentext{71.3}$^{\mathbf{**}}_{\pm0.7}$ / \greentext{72.9}$^{\mathbf{**}}_{\pm1.4}$ & 66.1$_{\pm 2.0}$ / 67.1$_{\pm 2.6}$ & 68.6$_{\pm 1.2}$ / 68.6$_{\pm 1.4}$ & \redtext{63.0}$_{\pm 2.5}$ / \redtext{63.9}$_{\pm 2.5}$ & 68.1$_{\pm 1.3}$ / 69.5$_{\pm 2.6}$ \\

        \midrule
        \multicolumn{6}{c}{\textbf{Gemini 1.5 Pro}} \\ % Changed version for clarity
        \midrule
        \textit{en}: \texttt{IMG} / \texttt{+CAPT}   & \greentext{70.9}$^{\mathbf{*}}_{\pm2.1}$ / \greentext{70.9}$^{\mathbf{*}}_{\pm2.1}$ & 69.7$_{\pm 2.1}$ / 69.7$_{\pm 2.1}$ & 68.6$_{\pm 1.0}$ / 68.6$_{\pm 1.0}$ & \redtext{65.0}$_{\pm 0.7}$ / \redtext{65.0}$_{\pm 0.7}$ & 66.7$_{\pm 2.7}$ / 66.7$_{\pm 2.7}$ \\
        \textit{de}: \texttt{IMG} / \texttt{+CAPT}   & 69.5$_{\pm 1.3}$ / \greentext{70.9}$_{\pm 1.7}$ & \greentext{70.7}$^{\mathbf{*}}_{\pm1.6}$ / 70.9$_{\pm 2.1}$ & 68.1$_{\pm 1.2}$ / 68.2$_{\pm 2.3}$ & \redtext{67.1}$_{\pm 2.2}$ / \redtext{66.8}$_{\pm 3.3}$ & 70.1$_{\pm 3.9}$ / 68.1$_{\pm 4.4}$ \\ 
        \textit{es}\kern+0.18em: \texttt{IMG} / \texttt{+CAPT}   & \greentext{69.5}$^{\mathbf{*}}_{\pm2.0}$ / \greentext{70.8}$^{\mathbf{*}}_{\pm3.2}$ & 69.2$_{\pm 1.7}$ / 68.8$_{\pm 4.3}$ & 68.7$_{\pm 1.5}$ / 66.7$_{\pm 3.1}$ & \redtext{65.4}$_{\pm 1.6}$ / \redtext{63.7}$_{\pm 2.8}$ & 69.0$_{\pm 2.8}$ / 65.9$_{\pm 5.0}$ \\ 
        \textit{hi}\kern+0.2em: \texttt{IMG} / \texttt{+CAPT}   & 61.4$_{\pm 5.1}$ / 63.5$_{\pm 3.1}$ & 61.4$_{\pm 8.2}$ / 65.4$_{\pm 4.4}$ & \greentext{63.9}$_{\pm 4.3}$ / 65.7$_{\pm 3.7}$ & 62.0$_{\pm 7.5}$ / \redtext{61.2}$_{\pm 4.5}$ & \redtext{57.8}$_{\pm 14.2}$ / \greentext{66.0}$_{\pm 6.7}$ \\ 
        \textit{zh}: \texttt{IMG} / \texttt{+CAPT}   & 60.8$_{\pm 2.5}$ / 63.7$_{\pm 4.6}$ & 62.6$_{\pm 4.2}$ / 63.4$_{\pm 6.0}$ & \greentext{66.0}$_{\pm 2.8}$ / \greentext{65.2}$_{\pm 5.4}$ & \redtext{60.4}$_{\pm 6.0}$ / \redtext{60.7}$_{\pm 6.6}$ & 63.1$_{\pm 6.3}$ / 62.8$_{\pm 7.4}$ \\

        \midrule
        \multicolumn{6}{c}{\textbf{Qwen2-VL 72B}} \\ % Changed version for clarity
        \midrule
        \textit{en}: \texttt{IMG} / \texttt{+CAPT}   & \greentext{71.5}$^{\mathbf{**}}_{\pm3.9}$ / \greentext{70.8}$^{\mathbf{*}}_{\pm4.8}$ & 62.3$_{\pm 3.5}$ / 62.4$_{\pm 3.9}$ & 65.5$_{\pm 3.7}$ / 65.4$_{\pm 3.9}$ & 59.1$_{\pm 3.9}$ / \redtext{58.0}$_{\pm 4.1}$ & \redtext{58.9}$_{\pm 4.4}$ / 58.2$_{\pm 4.7}$ \\
        \textit{de}: \texttt{IMG} / \texttt{+CAPT}   & \greentext{68.7}$^{\mathbf{**}}_{\pm0.8}$ / \greentext{70.1}$^{\mathbf{**}}_{\pm2.2}$ & 64.2$_{\pm 2.2}$ / 65.3$_{\pm 2.6}$ & 66.6$_{\pm 1.8}$ / 66.4$_{\pm 2.4}$ & \redtext{60.1}$_{\pm 1.4}$ / \redtext{59.2}$_{\pm 2.2}$ & 61.4$_{\pm 2.4}$ / 61.3$_{\pm 2.5}$ \\ 
        \textit{es}\kern+0.18em: \texttt{IMG} / \texttt{+CAPT}   & \greentext{70.8}$^{\mathbf{**}}_{\pm2.1}$ / \greentext{71.2}$^{\mathbf{**}}_{\pm2.3}$ & 62.5$_{\pm 2.2}$ / 63.4$_{\pm 3.1}$ & 65.4$_{\pm 1.5}$ / 66.1$_{\pm 2.6}$ & \redtext{59.3}$_{\pm 3.2}$ / 59.3$_{\pm 4.2}$ & 59.5$_{\pm 2.8}$ / \redtext{58.5}$_{\pm 3.2}$ \\ 
        \textit{hi}\kern+0.2em: \texttt{IMG} / \texttt{+CAPT}   & \greentext{62.9}$^{\mathbf{*}}_{\pm2.0}$ / \greentext{64.5}$^{\mathbf{*}}_{\pm3.2}$ & 58.2$_{\pm 3.6}$ / 58.4$_{\pm 4.0}$ & 61.7$_{\pm 4.3}$ / 61.8$_{\pm 5.2}$ & 55.8$_{\pm 3.2}$ / 56.1$_{\pm 3.0}$ & \redtext{54.9}$_{\pm 4.3}$ / \redtext{54.8}$_{\pm 3.8}$ \\ 
        \textit{zh}: \texttt{IMG} / \texttt{+CAPT}   & \greentext{66.1}$^{\mathbf{*}}_{\pm2.0}$ / \greentext{66.7}$^{\mathbf{*}}_{\pm2.2}$ & \redtext{58.3}$_{\pm 2.2}$ / 58.9$_{\pm 3.2}$ & 63.8$_{\pm 2.3}$ / 63.6$_{\pm 2.6}$ & 58.6$_{\pm 3.8}$ / \redtext{57.2}$_{\pm 4.2}$ & 60.6$_{\pm 3.8}$ / 59.9$_{\pm 4.4}$ \\

        \midrule
        \multicolumn{6}{c}{\textbf{LLaVA OneVision 73B}} \\ 
        \midrule      
        \textit{en}: \texttt{IMG} / \texttt{+CAPT}   & \greentext{71.2}$^{\mathbf{*}}_{\pm2.4}$ / \greentext{68.4}$^{\mathbf{**}}_{\pm1.4}$ & 69.1$_{\pm 2.1}$ / 61.6$_{\pm 2.2}$ & 69.3$_{\pm 2.7}$ / 62.9$_{\pm 1.8}$ & \redtext{64.3}$_{\pm 2.5}$ / \redtext{57.4}$_{\pm 1.5}$ & 66.2$_{\pm 2.8}$ / 58.2$_{\pm 2.0}$ \\
        \textit{de}: \texttt{IMG} / \texttt{+CAPT}   & \greentext{60.9}$^{\mathbf{*}}_{\pm1.5}$ / \greentext{65.6}$^{\mathbf{**}}_{\pm1.2}$ & 58.8$_{\pm 1.6}$ / 62.1$_{\pm 2.0}$ & 60.8$_{\pm 1.6}$ / 62.8$_{\pm 1.4}$ & \redtext{57.9}$_{\pm 2.1}$ / 59.0$_{\pm 1.4}$ & 59.5$_{\pm 2.2}$ / \redtext{57.6}$_{\pm 1.8}$ \\ 
        \textit{es}\kern+0.18em: \texttt{IMG} / \texttt{+CAPT}   & 62.9$_{\pm 1.0}$ / \greentext{64.8}$^{\mathbf{**}}_{\pm1.0}$ & 63.3$_{\pm 1.7}$ / 57.6$_{\pm 1.8}$ & \greentext{65.8}$^{\mathbf{**}}_{\pm1.5}$ / 59.4$_{\pm 1.4}$ & 59.8$_{\pm 1.2}$ / 55.8$_{\pm 2.6}$ & \redtext{57.8}$_{\pm 1.9}$ / \redtext{54.1}$_{\pm 1.6}$ \\ 
        \textit{hi}\kern+0.2em: \texttt{IMG} / \texttt{+CAPT}   & 58.2$_{\pm 1.4}$ / \greentext{64.1}$^{\mathbf{**}}_{\pm0.3}$ & 57.8$_{\pm 0.7}$ / 61.8$_{\pm 1.2}$ & \greentext{61.5}$^{\mathbf{**}}_{\pm1.2}$ / 63.3$_{\pm 0.1}$ & \redtext{52.3}$_{\pm 1.6}$ / 59.4$_{\pm 1.9}$ & 55.5$_{\pm 2.2}$ / \redtext{58.9}$_{\pm 1.9}$ \\ 
        \textit{zh}: \texttt{IMG} / \texttt{+CAPT}   & 55.7$_{\pm 0.7}$ / \greentext{65.3}$^{\mathbf{**}}_{\pm2.0}$ & 52.4$_{\pm 2.4}$ / 60.3$_{\pm 2.8}$ & \greentext{55.8}$^{\mathbf{*}}_{\pm1.4}$ / 60.2$_{\pm 2.2}$ & \redtext{49.1}$_{\pm 2.9}$ / 58.3$_{\pm 2.4}$ & 51.7$_{\pm 2.8}$ / \redtext{55.9}$_{\pm 3.0}$ \\
        
        \midrule
        \multicolumn{6}{c}{\textbf{InternVL2 76B}} \\ % Changed version for clarity
        \midrule
        \textit{en}: \texttt{IMG} / \texttt{+CAPT}   & \greentext{60.1}$^{\mathbf{*}}_{\pm3.0}$ / \greentext{65.1}$^{\mathbf{*}}_{\pm5.0}$ & 55.1$_{\pm 4.5}$ / 58.8$_{\pm 6.0}$ & 58.2$_{\pm 4.1}$ / 59.9$_{\pm 4.7}$ & 53.8$_{\pm 3.6}$ / 55.9$_{\pm 5.4}$ & \redtext{52.5}$_{\pm 4.6}$ / \redtext{54.1}$_{\pm 5.4}$ \\
        \textit{de}: \texttt{IMG} / \texttt{+CAPT}   & \greentext{57.1}$^{\mathbf{*}}_{\pm3.6}$ / \greentext{63.1}$^{\mathbf{*}}_{\pm3.8}$ & 52.6$_{\pm 5.4}$ / 57.7$_{\pm 6.3}$ & 54.0$_{\pm 5.3}$ / 58.8$_{\pm 4.3}$ & \redtext{50.8}$_{\pm 5.2}$ / 54.8$_{\pm 5.3}$ & 50.9$_{\pm 5.7}$ / \redtext{53.3}$_{\pm 5.4}$ \\ 
        \textit{es}\kern+0.18em: \texttt{IMG} / \texttt{+CAPT}   & \greentext{56.6}$^{\mathbf{*}}_{\pm3.3}$ / \greentext{62.1}$^{\mathbf{*}}_{\pm5.0}$ & 52.8$_{\pm 2.6}$ / 56.6$_{\pm 5.4}$ & 56.4$_{\pm 3.5}$ / 59.2$_{\pm 4.6}$ & 52.6$_{\pm 2.4}$ / 53.7$_{\pm 5.3}$ & \redtext{50.2}$_{\pm 3.3}$ / \redtext{51.8}$_{\pm 5.5}$ \\ 
        \textit{hi}\kern+0.2em: \texttt{IMG} / \texttt{+CAPT}   & \greentext{48.4}$^{\mathbf{*}}_{\pm1.8}$ / \greentext{59.1}$^{\mathbf{*}}_{\pm3.0}$ & 42.4$_{\pm 2.0}$ / 53.8$_{\pm 4.9}$ & 46.4$_{\pm 2.0}$ / 56.6$_{\pm 3.1}$ & 43.2$_{\pm 2.3}$ / 53.2$_{\pm 5.4}$ & \redtext{40.9}$_{\pm 2.9}$ / \redtext{49.8}$_{\pm 4.9}$ \\ 
        \textit{zh}: \texttt{IMG} / \texttt{+CAPT}   & \greentext{54.3}$_{\pm 2.1}$ / \greentext{59.4}$_{\pm 4.7}$ & 49.2$_{\pm 4.8}$ / 54.5$_{\pm 4.6}$ & 52.7$_{\pm 4.8}$ / 56.9$_{\pm 3.3}$ & 49.4$_{\pm 4.4}$ / 53.1$_{\pm 3.7}$ & \redtext{47.4}$_{\pm 6.5}$ / \redtext{52.2}$_{\pm 4.7}$ \\
        
        \bottomrule
    \end{tabular}
    \caption{The performance of our large VLMs across different meme languages while keeping the prompt in English. We report results using only the meme image as input (\texttt{IMG}) and also when including the image caption in the prompt (\texttt{+CAPT}). \greentext{Bold} text indicates the best performance across cultures; \redtext{underlined} text denotes the worst performance. An asterisk (\textbf{*}) indicates statistical significance compared to the lowest cultural performance, and a double asterisk (\textbf{**}) indicates significance compared to the second-highest cultural performance.}
    %Out of 50 variations, 42 achieve their best performance with US labels.}
    \label{tab:main}
\end{table*}

\paragraph{Annotators’ Disagreement Analysis} We conduct a qualitative analysis to examine \textit{why} cultures differ in their hate speech annotations, focusing on the pair with the highest disagreement: the USA and India. We recruit 7 annotators who are bilingual in Hindi and English, born in one of the two countries, currently residing in the other, and self-identifying as multicultural with ties to both cultures. These annotators are shown memes where the two cultures' annotations diverge, and we ask them to explain the reasons for their disagreement in free-text form. Using an inductive “bottom-up” approach, one author extracts keywords from each response, summarizing the text, giving us an initial codebook of 37 codes. A hired annotator then independently reassigns these established codes to the samples. We then establish 6 major themes.

As shown in Figure \ref{fig:pie_chart}, “Sensitivity Around Minority Groups” and “Social Norms \& Cultural Values” account for 53.6\%, while “Historical \& Political Context” and “Non-Existing Stereotypes” contribute 31\%. Together, these four themes, totaling 84.6\%, likely \textit{reflect cultural differences}. Ideally, we aim to minimize the proportion of “Language Error”, which accounts for only 5.2\%. However, 10.3\% fall under “Annotation Ambiguity”, which may stem from annotation noise or reflect individual annotators' personal preferences. In conclusion, our cross-cultural disagreements can largely be attributed to cultural differences.

\section{Experiments}

\subsection{Experimental Setup} \label{sec:exp}

\paragraph{Zero-Shot Setup} We evaluate VLMs using a zero-shot approach to detect hate speech. The task is framed as a multiple-choice format, where the model must select between two answers: (a) \textit{hate speech} and (b) \textit{non-hate speech}. We implement three different prompt variations, each altering the order of answers (a) and (b). In total, we generate six prompts, maintaining English as the prompts' language unless otherwise specified. Additionally, we experiment with two input variations: (1) using only the image (\texttt{IMG}) and (2) incorporating the image caption (\texttt{+CAPT}) into the prompt, see Appendix \ref{ap:prompt_var} for detailed prompts. 

\paragraph{Evaluation} We present the average accuracy across all prompt variations, along with the standard deviation. To determine whether the observed differences are statistically significant, we apply the Wilcoxon rank-sum test \cite{c4091bd3-d888-3152-8886-c284bf66a93a}, a non-parametric test that assesses whether one distribution tends to have higher values than another, without assuming normality.

\paragraph{Models} We evaluate several models, including GPT-4o\footnote{API Version: gpt-4o-2024-05-13} \cite{openai2024gpt4technicalreport}, Gemini 1.5 Pro\footnote{API Version: gemini-1.5-pro-001} \cite{geminiteam2024gemini15unlockingmultimodal}, Qwen2-VL \cite{wang2024qwen2vlenhancingvisionlanguagemodels}, LLaVA OneVision \cite{li2024llavaonevisioneasyvisualtask}, and InternVL2 \cite{chen2023internvl, chen2024far}. For more details, see Appendix \ref{ap:model_details}.

\subsection{Dataset Sanity Check}

We start by demonstrating the desired multimodality and evaluating the impact of different model scaling on our dataset. The aggregated results are presented in Table \ref{tab:sanity}, while detailed model performances in Table \ref{tab:unimodal} in the Appendix.

\paragraph{Multimodality} To demonstrate multimodality, we compare models that utilize images as input with those that rely solely on captions. We present the top-performing models for English input in both settings, based on average accuracy.

The top-performing multimodal model achieves an accuracy of 75.8\% with US labels, compared to 65.4\% for the best unimodal model. The significant higher accuracy of the multimodal models underscores the strength of our dataset in supporting multimodal analysis.

\paragraph{Scale} We compare the average performance of models within the same family, contrasting those with fewer than 10B parameters against those with more than 70B. On average, larger models exhibit better performance across all cultural labels, with the greatest improvement of 5.5\% seen on US labels. Our subsequent analysis focuses exclusively on large VLMs (models with over 70B parameters).

\subsection{Prompting in English} \label{sec:eng_prompt}

In this section, we report experiments with VLMs in a zero-shot setting, using English as the prompt language. The results are presented in Table \ref{tab:main}.

\paragraph{Strong Alignment with US Culture Label}
Across input and language variations, we observe that nearly all models perform best on US labels: out of 50 variations, 42 achieve their highest performance on US labels. In 39 cases, the performance difference compared to the worst-performing cultural label is statistically significant, and in 18 cases, the difference is significant even compared to the second-highest performing label.

For example, GPT-4o consistently performs best with US labels across all languages and input variants, achieving the highest accuracy on our dataset at 75.8\% for English. The model shows a significant difference from the second-highest cultural label in 8 out of 10 variations.

\paragraph{Low Alignment with Indian Culture Label} In contrast, the alignment between the model and hate speech annotations from Indian annotators is notably low, ranking among the bottom in accuracy across 30 out of 50 variants. Similarly, annotations from Chinese annotators also show low alignment, with 19 variants reflecting the lowest accuracy.

\paragraph{Comparison: \texttt{IMG} vs. \texttt{+CAPT}} Adding captions into the prompt improves performance in all languages except English, suggesting weaker OCR capabilities in VLMs for non-English text. For instance, LLaVA OneVision's accuracy on Hindi with US labels rises from 58.2\% to 64.5\% with captions.

\subsection{Prompting in Native Language} \label{sec:multilingual}

\begin{table}[t]
    \centering
    \small
    \setlength{\tabcolsep}{2.8pt}
    \begin{tabular}{l|ccccc}
        \toprule
         & \textbf{US} & \textbf{DE} & \textbf{MX} & \textbf{IN} & \textbf{CN} \\ 
        \midrule
        \multicolumn{6}{c}{\textbf{Multilingual Prompts: GPT-4o}} \\ % Changed version for clarity
        \midrule
        \textit{de}    & \textbf{75.0}$^{\mathbf{**}}_{\pm 1.0}$ & 71.6\textsubscript{$\pm 1.9$} & 69.4\textsubscript{$\pm 2.0$} & \underline{63.7}\textsubscript{$\pm 1.9$} & 67.2\textsubscript{$\pm 1.5$} \\
        - $\Delta$     & \tiny \textcolor{darkgreen}{+0.8} & \textcolor{darkgreen}{\textbf{+0.3 }} & \tiny \textcolor{darkgreen}{+0.4} & \tiny \textcolor{darkgreen}{+0.1} & \tiny \textcolor{darkgreen}{+0.5} \\[0.1cm] \hline

        \textit{es}   & \textbf{75.0}$^{\mathbf{**}}_{\pm 1.3}$ & 73.8\textsubscript{$\pm 1.1$} & 70.3\textsubscript{$\pm 1.8$} & \underline{64.1}\textsubscript{$\pm 1.1$} & 67.3\textsubscript{$\pm 2.6$} \\
        - $\Delta$     & \tiny \textcolor{red}{-0.6} & \tiny \textcolor{darkgreen}{+1.3} & \textcolor{red}{\textbf{-0.2 }} & \tiny \textcolor{darkgreen}{+0.4} & \tiny \textcolor{darkgreen}{+1.9} \\[0.1cm] \hline

        \textit{hi}     &  \textbf{72.8}$^{\mathbf{**}}_{\pm 0.9}$ & 70.0\textsubscript{$\pm 1.0$} & 71.2\textsubscript{$\pm 1.4$} & \underline{64.9}\textsubscript{$\pm 1.6$} & 67.4\textsubscript{$\pm 1.5$} \\
        - $\Delta$     & \tiny \textcolor{darkgreen}{+1.5\textbf{*}} & \tiny \textcolor{darkgreen}{+1.9\textbf{*}} & \tiny \textcolor{darkgreen}{+2.9\textbf{*}} & \textcolor{darkgreen}{\textbf{+1.0 }} & \tiny \textcolor{darkgreen}{+0.9} \\[0.1cm] \hline

        \textit{zh}  &  \textbf{72.4}$^{\mathbf{**}}_{\pm 1.3}$ & 66.4\textsubscript{$\pm 2.5$} & 69.3\textsubscript{$\pm 2.1$} & \underline{63.7}\textsubscript{$\pm 2.8$} & 70.2\textsubscript{$\pm 3.2$} \\
        - $\Delta$     & \tiny \textcolor{red}{-0.5} & \tiny \textcolor{red}{-0.7} & \tiny \textcolor{darkgreen}{+0.7} & \tiny \textcolor{red}{-0.2} & \textcolor{darkgreen}{\textbf{+0.7 }} \\

        \midrule
        \multicolumn{6}{c}{\textbf{Multilingual Prompts: Qwen2-VL 72B}} \\ % Changed version for clarity
        \midrule
        \textit{de}    & \textbf{69.9}$^{\mathbf{**}}_{\pm2.9}$ & 64.6$_{\pm 3.8}$ & 65.8$_{\pm 3.2}$ & \underline{58.8}$_{\pm 3.8}$ & 61.3$_{\pm 4.5}$ \\
        - $\Delta$     & \tiny \textcolor{red}{-0.2} & \textcolor{red}{\textbf{-0.7 }} & \tiny \textcolor{red}{-0.6} & \tiny \textcolor{darkgreen}{+0.4} & \tiny \textcolor{darkgreen}{+0.0} \\[0.1cm] \hline

        \textit{es}   &  \textbf{71.6}\textbf{*}\textsubscript{$\pm 3.1$} & 63.6\textsubscript{$\pm 3.4$} & 65.3\textsubscript{$\pm 3.0$} & 58.9\textsubscript{$\pm 3.7$} & \underline{57.8}\textsubscript{$\pm 4.1$} \\ 
        - $\Delta$     & \tiny \textcolor{darkgreen}{+0.4} & \tiny \textcolor{red}{-0.2} & \textcolor{red}{\textbf{-0.8 }} & \tiny \textcolor{red}{-0.4} & \tiny \textcolor{red}{-0.7} \\[0.1cm] \hline

        \textit{hi}     & \textbf{68.2}\textbf{*}\textsubscript{$\pm 2.7$} & 64.4\textsubscript{$\pm 4.9$} & 67.1\textsubscript{$\pm 4.6$} & \underline{61.2}\textsubscript{$\pm 5.0$} & 61.3\textsubscript{$\pm 6.2$} \\
        - $\Delta$     & \tiny \textcolor{darkgreen}{+3.7\textbf{*}} & \tiny \textcolor{darkgreen}{+6.1\textbf{*}} & \tiny \textcolor{darkgreen}{+5.3} & \textcolor{darkgreen}{\textbf{+5.1 }} & \tiny \textcolor{darkgreen}{+6.5} \\[0.1cm] \hline

        \textit{zh}  &  \textbf{67.1}\textbf{*}\textsubscript{$\pm 2.3$} & 62.7\textsubscript{$\pm 2.6$} & 65.8\textsubscript{$\pm 3.0$} & \underline{59.7}\textsubscript{$\pm 3.4$} & 61.6\textsubscript{$\pm 3.3$} \\
        - $\Delta$     & \tiny \textcolor{darkgreen}{+0.4} & \tiny \textcolor{darkgreen}{+3.8} & \tiny \textcolor{darkgreen}{+2.2} & \tiny \textcolor{darkgreen}{+2.5} & \textcolor{darkgreen}{\textbf{+1.7 }} \\

        \bottomrule
    \end{tabular}
    \caption{Evaluation of adjusting the prompt language to match the dominant language of the respective culture.  $\Delta$ shows the difference between the multilingual prompt (\texttt{+CAPT}) and English prompt (\texttt{+CAPT}). The asterisk (*) in the $\Delta$ row shows significant difference.} %There is no significant improvement on target culture.}
    \label{tab:multilingual}
\end{table}

\begin{table}[t]
    \centering
    \small
    \setlength{\tabcolsep}{2.8pt}
    \begin{tabular}{l|ccccc}
        \toprule
         & \textbf{US} & \textbf{DE} & \textbf{MX} & \textbf{IN} & \textbf{CN} \\ 
        \midrule
        \multicolumn{6}{c}{\textbf{Country Information: GPT-4o}} \\ % Changed version for clarity
        \midrule
        \textit{en}   & \textbf{74.3}$^{\mathbf{**}}_{\pm1.0}$ & 68.0$_{\pm 1.7}$ & 66.1$_{\pm 1.0}$ & \underline{60.1}$_{\pm 1.2}$ & 63.4$_{\pm 2.2}$ \\
        - $\Delta$     & \textcolor{red}{\textbf{-1.2 }} & \tiny \textcolor{red}{-3.7\textbf{*}} & \tiny \textcolor{red}{-3.2\textbf{*}} & \tiny \textcolor{red}{-3.1\textbf{*}} & \tiny\textcolor{red}{-4.2\textbf{*}} \\[0.1cm] \hline

        \textit{de}    & \textbf{73.5}$^{\mathbf{**}}_{\pm1.7}$ & 69.3$_{\pm 2.0}$ & 66.9$_{\pm 1.2}$ & \underline{62.7}$_{\pm 1.4}$ & 64.8$_{\pm 1.7}$ \\
        - $\Delta$     & \tiny\textcolor{red}{-0.7} & \textcolor{red}{\textbf{-2.0 }} & \tiny\textcolor{red}{-2.1\textbf{*}} & \tiny\textcolor{red}{-0.9} & \tiny\textcolor{red}{-2.4} \\[0.1cm] \hline
        
        \textit{es}   & \textbf{72.9}$^{\mathbf{**}}_{\pm1.4}$ & 70.2$_{\pm 1.4}$ & 67.7$_{\pm 1.7}$ & \underline{61.6}$_{\pm 2.2}$ & 64.2$_{\pm 2.5}$ \\
        - $\Delta$  & \tiny\textcolor{red}{-2.7\textbf{*}} & \tiny\textcolor{red}{-2.3\textbf{*}} & \textcolor{red}{\textbf{-2.8 }} & \tiny\textcolor{red}{-1.7} & \tiny\textcolor{red}{-1.2} \\[0.1cm] \hline

        \textit{hi}     & \textbf{67.8}$^{\mathbf{*}}_{\pm1.7}$ & 65.3$_{\pm 2.9}$ & 65.9$_{\pm 2.2}$ & \underline{61.2}$_{\pm 2.2}$ & 62.7$_{\pm 2.6}$ \\
        - $\Delta$ & \tiny\textcolor{red}{-3.5\textbf{*}} & \tiny\textcolor{red}{-2.8} & \tiny\textcolor{red}{-2.4} & \textcolor{red}{\textbf{-2.7 }} & \tiny\textcolor{red}{-3.8} \\[0.1cm] \hline

        \textit{zh}  & \textbf{69.9}$^{\mathbf{*}}_{\pm1.5}$ & 65.8$_{\pm 3.2}$ & 67.0$_{\pm 1.3}$ & \underline{62.4}$_{\pm 3.1}$ & 68.1$_{\pm 3.2}$ \\
        - $\Delta$  & \tiny\textcolor{red}{-3.0\textbf{*}} & \tiny\textcolor{red}{-1.3} & \tiny\textcolor{red}{-1.6} & \tiny\textcolor{red}{-1.5} & \textcolor{red}{\textbf{-1.4 }} \\

        \midrule
        \multicolumn{6}{c}{\textbf{Country Information: Qwen2-VL 72B}} \\ % Changed version for clarity
        \midrule
        \textit{en}   & \textbf{70.7}$^{\mathbf{**}}_{\pm4.3}$ & 61.6$_{\pm 3.6}$ & 64.1$_{\pm 2.6}$ & \underline{57.8}$_{\pm 4.4}$ & 58.4$_{\pm 4.3}$ \\
        - $\Delta$        & \textcolor{red}{\textbf{-0.1 }} & \tiny\textcolor{red}{-0.8} & \tiny\textcolor{red}{-1.3} & \tiny\textcolor{red}{-0.2} & \tiny\textcolor{darkgreen}{+0.2}  \\[0.1cm] \hline
        
        \textit{de}    & \textbf{68.5}$^{\mathbf{**}}_{\pm2.2}$ & 62.9$_{\pm 2.4}$ & 64.8$_{\pm 2.0}$ & \underline{56.8}$_{\pm 2.0}$ & 59.3$_{\pm 2.4}$ \\
        - $\Delta$  & \tiny\textcolor{red}{-1.6} & \textcolor{red}{\textbf{-2.4 }} & \tiny\textcolor{red}{-1.6} & \tiny\textcolor{red}{-2.4} & \tiny\textcolor{red}{-2.0} \\[0.1cm] \hline

        \textit{es}   & \textbf{69.9}$^{\mathbf{**}}_{\pm2.6}$ & 62.4$_{\pm 3.9}$ & 64.3$_{\pm 2.3}$ & 58.3$_{\pm 3.0}$ & \underline{57.6}$_{\pm 3.6}$ \\
        - $\Delta$       & \tiny\textcolor{red}{-1.3} & \tiny\textcolor{red}{-1.0} & \textcolor{red}{\textbf{-1.8 }} & \tiny\textcolor{red}{-1.0} & \tiny\textcolor{red}{-0.9} \\[0.1cm] \hline

        \textit{hi}     & \textbf{62.9}$^{\mathbf{*}}_{\pm2.3}$ & 58.7$_{\pm 4.0}$ & 62.2$_{\pm 3.9}$ & 56.1$_{\pm 3.2}$ & \underline{54.7}$_{\pm 3.7}$ \\
        - $\Delta$ & \tiny\textcolor{red}{-1.6} & \tiny\textcolor{darkgreen}{+0.3} & \tiny\textcolor{darkgreen}{+0.4} & \textcolor{darkgreen}{\textbf{+0.0 }} & \tiny\textcolor{red}{-0.1} \\[0.1cm] \hline

        \textit{zh}  & \textbf{63.9}$^{\mathbf{*}}_{\pm2.0}$ & 58.1$_{\pm 2.7}$ & 61.9$_{\pm 2.4}$ & \underline{54.3}$_{\pm 3.1}$ & 57.2$_{\pm 3.4}$ \\
        - $\Delta$  & \tiny\textcolor{red}{-2.8} & \tiny\textcolor{red}{-0.8} & \tiny\textcolor{red}{-1.7} & \tiny\textcolor{red}{-2.9} & \textcolor{red}{\textbf{-2.7 }}\\
        
        \bottomrule
    \end{tabular}
    \caption{Evaluation of injecting the country information. $\Delta$ represents the difference between the prompt with country information injection and those without it. The asterisk (*) in the $\Delta$ row shows significant difference.} 
    %There is no significant improvement.}
    \label{tab:country}
\end{table}

We experiment with adjusting the prompt language to match the dominant language of the target culture, as shown in Figure \ref{fig:multilingual_prompt} in the Appendix. Incorporating captions in the prompt (\texttt{+CAPT}) typically enhances performance, so we focus on this setting with the best-performing open-source and proprietary models. Results are presented in Table \ref{tab:multilingual}

%\paragraph{No Significant Improvement on Target Culture} 
The results show mixed effects: e.g., switching to German improves the performance of GPT-4o by 0.3, while for Qwen2, it decreases by 0.7. However, none of the observed changes are statistically significant. Therefore, we conclude that \textbf{altering the prompt language to match the dominant language of a specific culture does not have a meaningful impact on aligning models}.

%\paragraph{Strong Alignment with US Culture Label} 
Even when the prompt's language is altered, the model continues to show high alignment  with US labels. All variations significantly outperform the lowest-performing cultural variant with US culture label, and, for GPT-4o, they even significantly surpass the second-highest. This reinforces the idea that the \textbf{models are more aligned with US cultural norms, even when prompted in the dominant language of another culture}.

\subsection{Adding Country Information} \label{sec:country}

Building on \citet{lee-etal-2024-exploring-cross}, we align VLMs with the target culture by adding country information to the prompt. We report results only for the \texttt{+CAPT} setup and best models, as shown in Table \ref{tab:country}.

%\paragraph{No Significant Improvement} 
Injecting country information generally decreases performance across target cultures, with the exception of Qwen2 in Hindi, which shows no change. For instance, adding ``Germany'' to the prompt results in a 2.0 and 2.4-point accuracy drop for GPT-4o and Qwen2-VL, respectively, though these decreases are not statistically significant. Therefore, we conclude that \textbf{adding country information does not positively impact performance in the target culture}.

\section{Conclusion}

We present the first multimodal, parallel, multilingual hate speech dataset, annotated by a multicultural set of annotators. This dataset contains 300 parallel meme samples across five languages and has been annotated for hate speech across five cultures. We show that cultural factors significantly impact multimodal hate speech annotations in our dataset. Additionally, we use this dataset to highlight that VLMs exhibit a strong cultural bias towards the US, independent on the image and prompt language.

\section*{Limitation} \label{sec:Limitation}

While our dataset contains 300 samples across 5 languages—amounting to 1500 memes in total—the relatively small size reflects the challenges of generating high-quality translations and culturally diverse annotations. Expanding such a dataset is resource-intensive, both in terms of cost and labor.

Additionally, the dataset was sourced from a single website, primarily in English, and does not specifically target content from various cultural contexts. Furthermore, by selecting annotators from Prolific and using a single language per country, we introduce a degree of selection bias, as this method may not fully represent the complex cultural landscapes within each country.

We also recognize that equating culture with country is a limitation, as countries are often multicultural and multiethnic. For example, India is home to thousands of ethnic and tribal groups \cite{thapar2024india}, and our approach does not fully capture this diversity.

Finally, while our work highlights cross-cultural differences in the perception of hate speech, understanding the root causes of these disagreements remains an open challenge. Although we offer a qualitative analysis of annotator disagreements, a comprehensive theory-driven analysis is still lacking. Developing a robust theoretical framework to explain these cultural variations could ultimately help the alignment of VLMs with specific cultural nuances, leading to more accurate and culturally sensitive hate speech detection systems.

\section*{Ethics Statement} \label{sec:ethics}

The annotators recruited through Prolific were compensated at a rate of £10.65 per hour, in alignment with the minimum wage in the authors' country, ensuring fair payment. Prior to the start of annotations, the project received ethical approval from the lead author’s institution. All annotators were thoroughly informed about the nature of the project, including warnings regarding potentially harmful and offensive content. Each annotator provided explicit consent before beginning their work, ensuring they were fully aware of the content and the purpose of their involvement.

We also acknowledge the potential risks associated with distributing our dataset. To mitigate these risks, we will establish clear terms of use that strictly prohibit any form of malicious exploitation. Additionally, we release the dataset in an anonymized format, ensuring that all user IDs and any personally identifiable information are removed to protect individual privacy.

We use AI assistants, specifically GPT-4o, to help edit sentences in our paper writing. \name{} is licensed under CC BY-NC-ND 4.0.

\section*{Acknowledgement}

The work of Minh Duc Bui and Katharina von der Wense is funded by the Carl Zeiss Foundation, grant number P2021-02-014 (TOPML project). The work of Anne Lauscher is funded under the Excellence Strategy of the German Federal Government and the Federal States. 
We thank Sukannya Purkayastha, Pranav A, Yujie Ren, Zhu Luan, Timm Dill, Carlos Galarza, and Delia Rieger for helping with translations and feedback on non-English text. We also thank Kyung Eun Park, Carolin Holtermann and Abteen Ebrahimi for their helpful feedback and discussions.

\bibliography{acl_latex}

\appendix

\section{Dataset Construction Details}

\subsection{Topic List} \label{ap:topics}
We explain, how we further divide each sociopolitical category into smaller topics:
(1) \textbf{Religion}, divided into the world’s major religions \cite{religion}; (2)~\textbf{Nationalities}, aligned with our target countries; (3)~\textbf{Ethnicity}, structured as outlined in \citet{cheng-etal-2023-marked}; (4) \textbf{LGBTQ+}, representing the groups denoted by the acronym; and (5) \textbf{Political Issues}, identified as cultural issues during the US election, as defined by \citet{pew2024}.

\subsection{Keyword Matching} \label{ap:keyword_matching}

Following keyword matching, we retain only the templates with at least 10 captions (after pre-filtering). Additionally, each topic must have a minimum of three templates that meet this criterion to ensure a diverse set of templates per topic. Topics that do not meet these requirements are filtered out.

\begin{table}[t]
    \centering
    \tiny
    \setlength{\tabcolsep}{5pt}
    \begin{tabular}{l|ccccc}
    \toprule
    & USA & Germany & Mexico & India & China  \\
    \midrule
    \textbf{No. of Annotators} & 105 & 103 & 101 & 66 & 70 \\ \midrule
    \textbf{Gender} (\%) &  &  &  &  &  \\
    \quad male & 53.33 & 49.51 & 52.48 & 53.03 & 48.57 \\ 
    \quad female & 46.67 & 50.49 & 47.52 & 46.97 & 51.43 \\
    \quad non-binary & -- & -- & -- & -- & -- \\ \midrule
    \textbf{Ethnicity (Simplified)} (\%) &  &  &  &  &  \\
    \quad Asian & 1.9 & -- & -- & 93.94 & 97.14 \\ 
    \quad Black & 15.24 & -- & -- & -- & -- \\ 
    \quad White & 82.86 & 100.0 & 28.71 & -- & 1.43 \\ 
    \quad Mixed & -- & -- & 41.58 & -- & 1.43 \\ 
    \quad Other & -- & -- & 29.7 & 6.06 & -- \\ \midrule
    \textbf{Level of Education} (\%) &  &  &  &  &  \\ 
    \quad Below High School & -- & -- & 0.99 & -- & -- \\ 
    \quad High School & 24.76 & 22.33 & 9.9 & 3.03 & -- \\ 
    \quad College & 20.95 & 16.5 & 17.82 & 3.03 & 7.14 \\ 
    \quad Bachelor & 37.14 & 38.83 & 60.4 & 71.21 & 32.86 \\ 
    \quad Master’s Degree & 14.29 & 19.42 & 9.9 & 22.73 & 48.57 \\ 
    \quad Doctorate & 2.86 & 2.91 & 0.99 & -- & 11.43 \\ \midrule
    \textbf{Age} (\%) &  &  &  &  &  \\ 
    \quad 18-19 & 1.9 & 4.85 & 1.98 & 9.09 & 1.43 \\ 
    \quad 20-29 & 28.57 & 64.08 & 74.26 & 60.61 & 60.0 \\ 
    \quad 30-39 & 35.24 & 21.36 & 22.77 & 16.67 & 30.0 \\ 
    \quad 40-49 & 17.14 & 8.74 & -- & 9.09 & 5.71 \\ 
    \quad 50-59 & 13.33 & 0.97 & 0.99 & 4.55 & 1.43 \\ 
    \quad 60-69 & 3.81 & -- & -- & -- & 1.43 \\ 
    \quad 70-79 & -- & -- & -- & -- & -- \\ 
    \quad 80-89 & -- & -- & -- & -- & -- \\ \midrule
    \textbf{Political Orientation} (\%) &  &  &  &  &  \\ 
    \quad Liberal/Progressive & 35.24 & 37.86 & 34.65 & 25.76 & 8.57 \\ 
    \quad Moderate Liberal & 20.95 & 32.04 & 26.73 & 18.18 & 22.86 \\ 
    \quad Independent & 22.86 & 15.53 & 13.86 & 43.94 & 37.14 \\ 
    \quad Moderate Conservative & 10.48 & 12.62 & 14.85 & 9.09 & 12.86 \\ 
    \quad Conservative & 8.57 & 0.97 & 3.96 & 1.52 & -- \\
    \quad Other & 1.9 & 0.97 & 5.94 & 1.52 & 18.57 \\ \midrule
    \textbf{Religion} (\%) &  &  &  &  &  \\ 
    \quad None & 42.86 & 70.87 & 42.57 & -- & 74.29 \\ 
    \quad Christian & 46.67 & 28.16 & 37.62 & 6.06 & 5.71 \\ 
    \quad Buddhism & -- & -- & -- & 3.03 & 8.57 \\ 
    \quad Islam & 1.9 & -- & -- & 21.21 & -- \\ 
    \quad Judaism & -- & -- & -- & -- & -- \\ 
    \quad Hinduism & -- & -- & -- & 60.61 & -- \\ 
    \quad Irreligion & 2.86 & 0.97 & 8.91 & 1.52 & 2.86 \\ 
    \quad Other & 5.71 & -- & 10.89 & 7.58 & 8.57 \\ \midrule
    \bottomrule
    \end{tabular}
    \caption{Demographics of annotators during the hate speech annotation phase.}
    \label{table:demographic}
\end{table}

\subsection{Pre-Filtering} \label{ap:pre-filtering}

\subsubsection{Filtering for English Captions}

To filter out non-English captions, we utilize the implementation by \citet{nakatani2010langdetect}, which employs a Naive Bayes approach based on n-grams \cite{cavnar}.

\subsubsection{Filtering for Multimodal Hate Speech} \label{ap:multimodal}

We outline our method for filtering potentially multimodal hate speech samples by comparing two types of classifications: (1) user captions combined with manually created image descriptions and (2) the user captions alone. By comparing the outcomes of these two classifications, we identify content as multimodal hate speech when the first case (caption + image description) is flagged as hate speech, but the second case (caption only) is not.

\paragraph{Experimental Setup}
We employ zero-shot learning with Llama 3 \cite{dubey2024llama3herdmodels} to detect hate speech. For the image descriptions, two annotators were tasked with generating descriptions for all meme templates. A sample is only classified as hate speech in scenario (1) if both image descriptions+caption are classified as hate speech by the model. The prompt used is reported in Figure \ref{fig:prompt_classifcation}.
\begin{figure}
    \centering
    \begin{tcolorbox}
    \begin{Verbatim}[fontsize=\small]
    Classify the following meme into 
    'hatespeech' or 'no hatespeech'. 
    The definition of hatespeech is:
    <definition>
    Be culturally aware and sensitive.
    Meme Image: <image description>
    Text: <caption>
    \end{Verbatim}
    \end{tcolorbox}
    \caption{Prompt for classification of multimodal hate speech in the pre-filtering stage.}
    \label{fig:prompt_classifcation}
\end{figure}

\subsubsection{Filtering for Wordplay}

To ensure captions are easily translatable and avoid noise from wordplay, two fluent English speakers classify each caption as either non-wordplay or wordplay. Only captions unanimously classified as non-wordplay are selected.

\subsection{Translation Stage Details} \label{ap:translation}
To translate the captions by the Google Translate API, each caption is inputted with a separator `` // '' to clearly distinguish between the top and bottom text.

Furthermore, each human translator is provided with a detailed annotation guide outlining the criteria for what constitutes a correct translation and how corrections should be made. The annotation guidelines are shown in Figure \ref{fig:translation_guide}.

\begin{figure}[t]
    \centering
\begin{tcolorbox}
\begin{Verbatim}[fontsize=\small]
Correct Translation (1):
- The translation should accurately convey 
  the same meaning as the original text.
- It does not have to be grammatically 
  perfect (memes are rarely grammatically 
  correct), but it should be easily 
  understandable in the target language.
- Ensure that the ``<sep>'' marker is placed 
  in a position that maintains the same 
  semantic meaning as in the original 
  context (if possible).
- It should maintain the appropriate 
  context and style for the given content.
- Proper cultural references and idiomatic 
  expressions should be correctly translated 
  or adapted as needed.

False Translation (0):
- The translation significantly distorts or 
  changes the meaning of the original text.
- It contains major grammatical errors that 
  hinder understanding.
- It fails to convey the intended context, 
  tone, or style of the original text.
- Key terms or names are mistranslated or 
  omitted.
- Important nuances or details from the 
  original text are lost or incorrectly 
  translated.
\end{Verbatim}
    \end{tcolorbox}
    \caption{Annotation guidelines for translators.}
    \label{fig:translation_guide}
\end{figure}

\subsection{Pre-Annotation} \label{ap:preannotation}

Before beginning the main annotation process, we conduct a pre-annotation stage to balance the dataset. For this phase, we create a parallel multilingual meme dataset consisting of 450 samples. The dataset is divided into five equal parts, with each part assigned to annotators from a different cultural background. Each sample is annotated twice, with a total of 50 annotators involved—10 from each cultural group.

To achieve balance, we adjust the dataset so that 40\% of the samples are labeled as hate speech, 40\% as non-hate speech, and 20\% where there was a tie between annotators. This process results in the final set of 300 samples.

\begin{figure}[t]
    \centering
    \includegraphics[width=0.8\linewidth]{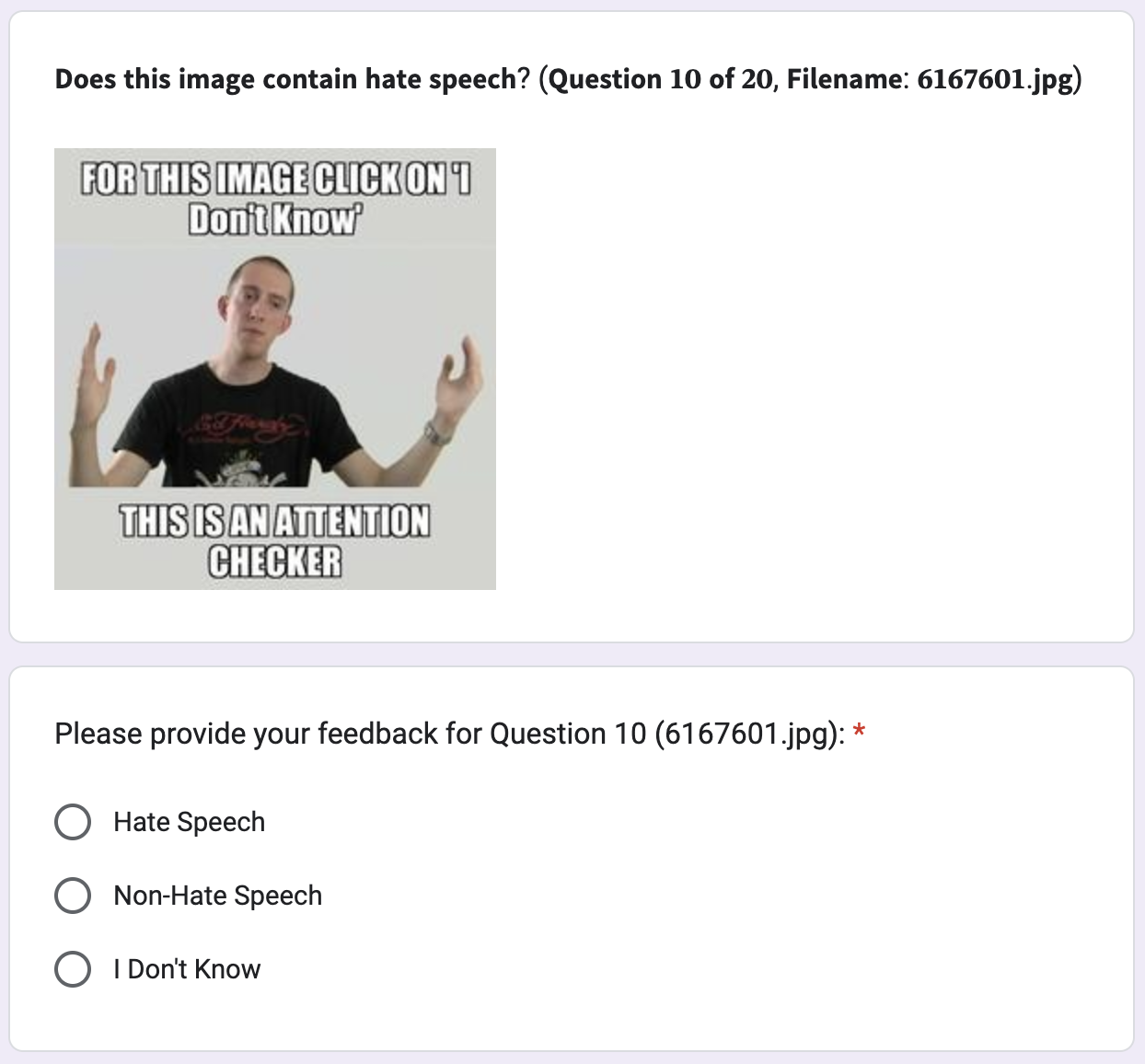}
    \caption{An example of an explicit attention check used in our survey.}
    \label{fig:explicit}
\end{figure}

\subsection{Hate Speech Survey Design} \label{ap:survey_design}

We use Google Forms\footnote{\url{https://www.google.de/intl/de/forms/about}} to design and distribute our surveys. To create surveys in each target language, we use the v3 Google Translate API to translate the surveys, which are then reviewed and corrected by native speakers for accuracy. We then create fixed random parallel batches, which are then assigned to each annotator. 

Each batch includes four attention checks: one explicit check, where annotators are required to select a specific pre-defined answer (see Figure \ref{fig:explicit}), and three implicit checks. These implicit checks consist of samples presented as examples at the beginning of the survey, accompanied by explanations of why the samples are classified as non-hate speech or hate speech based on the given definition.

We only retain annotations where the explicit attention check is answered correctly, and at least two out of the three implicit checks are passed. After collecting five annotations per sample, we review the results for any ties that need resolution and create new batches accordingly.

\begin{figure}[t]
    \centering
\begin{tcolorbox}
\begin{Verbatim}[fontsize=\small]
<PREFIX>:
Definition of Hate Speech:

Any kind of communication [...], that 
attacks or uses pejorative or discriminatory 
language with reference to a person or a 
group on the basis of who they are, in other
words, based on their religion, ethnicity, 
nationality, race, colour, descent, gender 
or other identity factor.
\end{Verbatim}
\end{tcolorbox}
\begin{tcolorbox}
\begin{Verbatim}[fontsize=\small]
<PROMPT #1>:
Answer if this meme is hate or not (in 
<country>) with a single alphabet letter 
among given answer choices a and b.
\end{Verbatim}
\end{tcolorbox}
\begin{tcolorbox}
\begin{Verbatim}[fontsize=\small]
<PROMPT #2>:
Given a meme from online, answer if it is 
'Hate Speech' or 'Non-Hate Speech' (in 
<country>). Answer with only one single
alphabet between 'a' and 'b', with no other
description.
\end{Verbatim}
\end{tcolorbox}
\begin{tcolorbox}
\begin{Verbatim}[fontsize=\small]
<PROMPT #3>:
Would you consider this meme as a hate  
speech (in <country>)? Give an answer in a 
single alphabet only.
\end{Verbatim}
\end{tcolorbox}
\begin{tcolorbox}
\begin{Verbatim}[fontsize=\small]
<POSTFIX>:
Meme: <image>
(Caption inside the meme image: <caption>)
Choose:
a: Hate
b: Non-hate
\end{Verbatim}
\end{tcolorbox}
    \caption{All three prompt variations: The order of options (a) and (b) is switched to create a total of six variations. Brackets are optional, allowing for insertion of the caption (\texttt{+CAPT} Setting) or country information as described in Section \ref{sec:country}.}
    \label{fig:prompt_variations}
\end{figure}

\subsection{Terms of Use}

Our research is conducted in the public interest under the GDPR, fulfilling the conditions for substantial public interest as academic research. We were unable to locate any Terms of Service on \url{https://memegenerator.net}, and the contact information provided on the website appears to be outdated and non-functional. To ensure we respect the platform's rights, we are publishing \name{} under the CC BY-NC-ND 4.0 license.

\subsection{Time Required for Dataset Development}

Estimating the effort required to create such a dataset is challenging due to the multiple, often unforeseen, refinement stages involved. For instance, unexpected challenges—such as translating wordplay—necessitated filtering them out and revisiting the translation process, significantly increasing the time and effort required. Overall, the entire dataset creation process took approximately four months from start to finish.

\begin{table*}[t]
    \centering
    \small
    \begin{tabular}{l|ccccc}
        \toprule
        \setlength{\tabcolsep}{2pt}

        \diagbox{\textbf{Inp.}}{\textbf{GT}} & \textbf{US} & \textbf{DE} & \textbf{MX} & \textbf{IN} & \textbf{CN} \\ 
        \midrule
        \multicolumn{6}{c}{\textbf{Unimodal}} \\ % Changed version for clarity

        \midrule
        GPT-4o & \textbf{65.4}$_{\pm3.9}$ & 58.3$_{\pm 4.4}$ & 60.8$_{\pm 3.5}$ & 56.7$_{\pm 4.0}$ & 55.0$_{\pm 4.5}$ \\
        Gemini 1.5 Pro  & 63.5$_{\pm3.9}$ & \textbf{59.2}$_{\pm 5.5}$ & 60.1$_{\pm 5.2}$ & 56.8$_{\pm 5.1}$ & 54.6$_{\pm 4.4}$ \\
        InternVL2 76B & 59.2$_{\pm3.1}$ & 52.1$_{\pm 3.0}$ & \textbf{60.8}$_{\pm 2.6}$ & \textbf{56.8}$_{\pm 4.1}$ & \textbf{55.0}$_{\pm 3.4}$ \\
        LLaVA OneVision 73B & 61.8$_{\pm1.5}$ & 54.3$_{\pm 1.3}$ & 56.5$_{\pm 1.1}$ & 51.7$_{\pm 1.1}$ & 49.7$_{\pm 1.8}$ \\
        Qwen2-VL 72B   & 61.9$_{\pm2.5}$ & 53.7$_{\pm 2.9}$ & 56.2$_{\pm 2.1}$ & 51.5$_{\pm 2.8}$ & 49.4$_{\pm 2.9}$ \\
        \midrule
        \multicolumn{6}{c}{\textbf{$\mathbf{<10}$B Models}} \\ 
        \midrule
        Qwen2-VL 7B   & \textbf{67.4}$^{\mathbf{*}}_{\pm 2.0}$ & \textbf{67.2}$^{\mathbf{*}}_{\pm 4.1}$ & \textbf{68.2}$^{\mathbf{*}}_{\pm2.6}$ & \textbf{62.3}$^{\mathbf{*}}_{\pm 3.4}$ & \textbf{65.3}$^{\mathbf{*}}_{\pm 3.9}$ \\

        InternVL2 8B & 58.2$_{\pm 2.8}$ & 57.4$_{\pm 4.0}$ & 58.7$_{\pm 4.9}$ & 57.6$_{\pm 4.8}$ & \underline{56.7}$_{\pm 4.2}$ \\
        Llava OneVision 7B & 60.3$_{\pm 5.5}$ & 55.6$_{\pm 7.3}$ & 58.3$_{\pm 6.9}$ & 54.1$_{\pm 6.9}$ & 52.3$_{\pm 7.4}$ \\
        \bottomrule
    \end{tabular}
    \caption{Unimodal setting: Models only get the caption as the input. The best value in each column is bolded. Sorted by average accuracy.}
    \label{tab:unimodal}
\end{table*}

\section{Experiments Details}

\subsection{Model Details} \label{ap:model_details}

Table \ref{table:model_list} lists the models and their sizes, all run on three H100 GPUs. Each large VLM processes five languages in around 1.5 hours. To support better text extraction from memes, images are resized to 512x512 pixels \cite{beyer2024vitspeed}. For all models, we generate deterministic outputs and limit generation to 40 new tokens. For the Gemini 1.5 Pro model, we disable all safety settings to minimize rejected responses. 

To derive binary classifications from the answers, we implement a custom keyword extraction. We relax the constraints on possible answers significantly, moving beyond a binary choice of “a” or “b”. For instance, “non-hate” or “hate-speech” is also recognized as a valid response in our analysis. However, answers that are nonsensical are counted as incorrect.

\subsection{Prompts} \label{ap:prompt_var}
\begin{figure*}[t]
    \centering
    \includegraphics[width=1\linewidth]{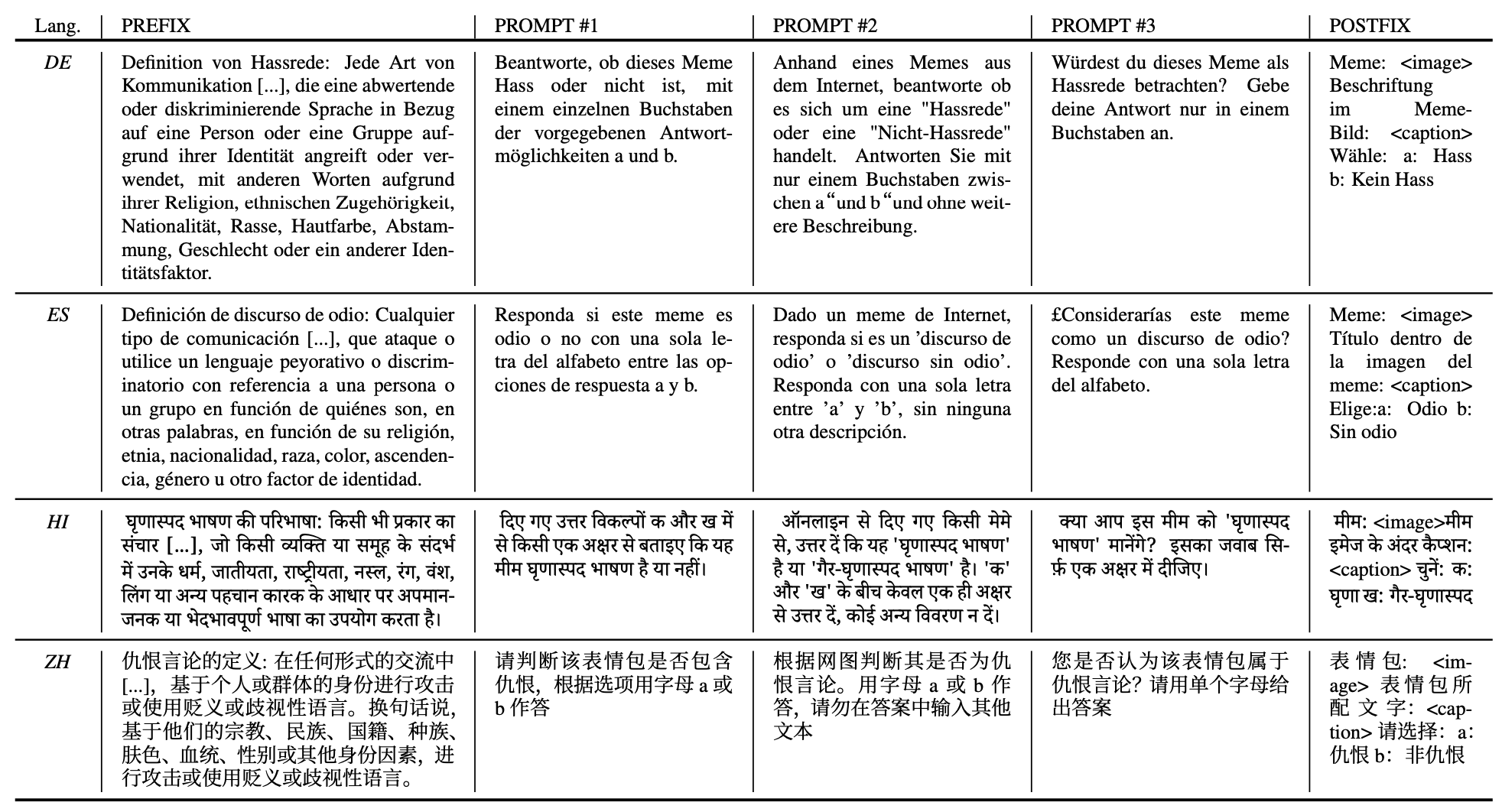}
    \caption{We present all multilingual prompts after removing any spaces, which correspond analogously to the English version shown in Figure \ref{fig:prompt_variations}.}
    \label{fig:multilingual_prompt}
\end{figure*}

We design prompts similar to those in \citet{lee-etal-2024-exploring-cross} and present the various prompt formulations in Figure \ref{fig:prompt_variations}. Additionally, the multilingual version of these prompts is shown in Figure \ref{fig:multilingual_prompt}.

\section{Supervised Baseline}

We experiment with a small supervised VLM baseline, training the Qwen2-VL 7B separately on each cultural label (i.e., culture-specific models) on only one prompt variation, using a 3-fold cross-validation approach due to the relatively small dataset size. We fine-tune our model using LoRA \cite{hu2021loralowrankadaptationlarge}, modifying only the Query and Value matrices, with a rank of 8 and an alpha value of 16. We employ a learning rate of $2e-4$ with a constant learning rate schedule, training for three epochs with a batch size of 16. Additionally, we refrain from hyperparameter tuning to avoid overfitting on our validation folds, as our limited data prevents the creation of a separate test set. The results are presented in Table \ref{tab:supervised}.

Overall, we observe high improvements across all countries, with the most notable gain of 7.1 points in Indian annotations. Moreover, the highest level of agreement with each culture-specific annotation is achieved by the model fine-tuned on the respective cultural labels. This suggests that each country's cultural perception can be effectively improved through supervised fine-tuning. We leave a more thorough analysis of the results and the exploration of advanced fine-tuning approaches for future work.

\begin{table}[t]
    \centering
    \small
    \begin{tabular}{l|ccccc}
        \toprule
        \setlength{\tabcolsep}{2pt}

        \diagbox{\textbf{Inp.}}{\textbf{GT}} & \textbf{US} & \textbf{DE} & \textbf{MX} & \textbf{IN} & \textbf{CN} \\ 
        \midrule
        Zero-Shot  & 67.4 & 67.2 & 68.2 & 62.3 & 65.3 \\ \midrule
        Fine-Tuned - US   & \textbf{71.5} & 69.8 & 70.9 & 65.9 & 66.9 \\
        Fine-Tuned - DE & 70.1 & \textbf{74.1} & 71.9 & 67.5 & 70.9 \\
        Fine-Tuned - MX   & 69.2 & 73.2 & \textbf{72.3} & 68.1 & 70.4 \\
        Fine-Tuned - IN   & 69.5 & 72.1 & 71.0 & \textbf{69.4} & 71.8 \\
        Fine-Tuned - CN   & 70.1 & 72.3 & 70.2 & 68.8 & \textbf{71.9} \\
        \bottomrule
    \end{tabular}
    \caption{We compare the zero-shot performance of Qwen2-VL 7B with a version fine-tuned separately for each cultural label using supervised learning and 3-fold cross-validation. We bold the best performance on each cultural label.}
    \label{tab:supervised}
\end{table}

\begin{table}[t]
    \small
    \centering
    \begin{tabular}{@{}l|r}
    \toprule
    \textbf{Model} & \textbf{|Total|} \\
    \midrule
    \href{https://huggingface.co/llava-hf/llava-onevision-qwen2-7b-ov-hf}{LLaVA-Onevision 7B} & $8.03\mathrm{B}$ \\    \href{https://huggingface.co/llava-hf/llava-onevision-qwen2-72b-ov-hf}{LLaVA-Onevision 72B} & $73.2\mathrm{B}$ \\ \hline

    \href{https://huggingface.co/Qwen/Qwen2-VL-7B-Instruct}{Qwen2-VL-7B-Instruct} & $8.29\mathrm{B}$ \\    \href{https://huggingface.co/Qwen/Qwen2-VL-72B-Instruct}{Qwen2-VL-72B-Instruct} & $73.4\mathrm{B}$ \\ \hline

    \href{https://huggingface.co/OpenGVLab/InternVL2-8B}{InternVL2-8B} & $8.08\mathrm{B}$ \\     
    \href{https://huggingface.co/OpenGVLab/InternVL2-Llama3-76B}{InternVL2-Llama3-76B} & $76.3\mathrm{B}$ \\ \hline
    Gemini 1.5 Pro & ? \\ \hline
    GPT-4o & ? \\

    \bottomrule
    \end{tabular}
    \caption{We present the models used in this study, along with their respective total number of parameters (denoted as ``|Total|''). Each model name is hyperlinked to its corresponding Huggingface repository (when viewed digitally). For Gemini 1.5 Pro and GPT-4o, we use gemini-1.5-pro-001 and gpt-4o-2024-05-13, respectively.}
    \label{table:model_list}
\end{table}

\begin{table}[t]
    \centering
    \small
    \begin{tabular}{c|c|c}
    \toprule
       \textbf{Topic} & \textbf{Keywords} & \textbf{Count}\\ \midrule
         Christianity & christ, jesus, priest & 21 \\
         Islam & muslim, islam & 22  \\
         Hinduism & hindu, hinduism & --  \\
         Buddhism & buddha, buddhist & --  \\
         Folk Religion & folk religion & --  \\
         Judaism & jew, judaism & 18  \\ \midrule
         Germany & germany, german & 18  \\ 
         United States & america, usa, american & 21\\ 
         Mexico & mexico, mexcian & 20\\ 
         China & china, chinese & 21\\ 
         India & india, indian & 15\\ \midrule         
         Asian & asia, asien & 20\\ 
         Black & black & 23\\ 
         Latine & latino, latine & --\\ 
         Middle Eastern & middle+eastern, arab & 19\\ 
         White & white & 19\\ \midrule
         Lesbian & lesbian & --\\
         Gay & gay & --\\
         Bisexual & bisexual & --\\
         Transgender & trans, transgender & 19\\ 
         Queer & queer & --\\ \midrule
         Law Enforcement & police & 23 \\ 
         Feminism & feminist & 21\\ 
         Immigration & immigrants & --\\ 
         Racial Diversity & (already included) & --\\ 
         LGBTQ+ & (already included) & --\\ 
        \bottomrule
    \end{tabular}
    \caption{We conduct a keyword search based on the identified topics and report the final sample count in \name. A ``--'' indicates that the topic did not meet our requirements.}
    \label{tab:keywords}
\end{table}

\begin{figure*}[t]
    \centering
    \includegraphics[width=0.36\linewidth]{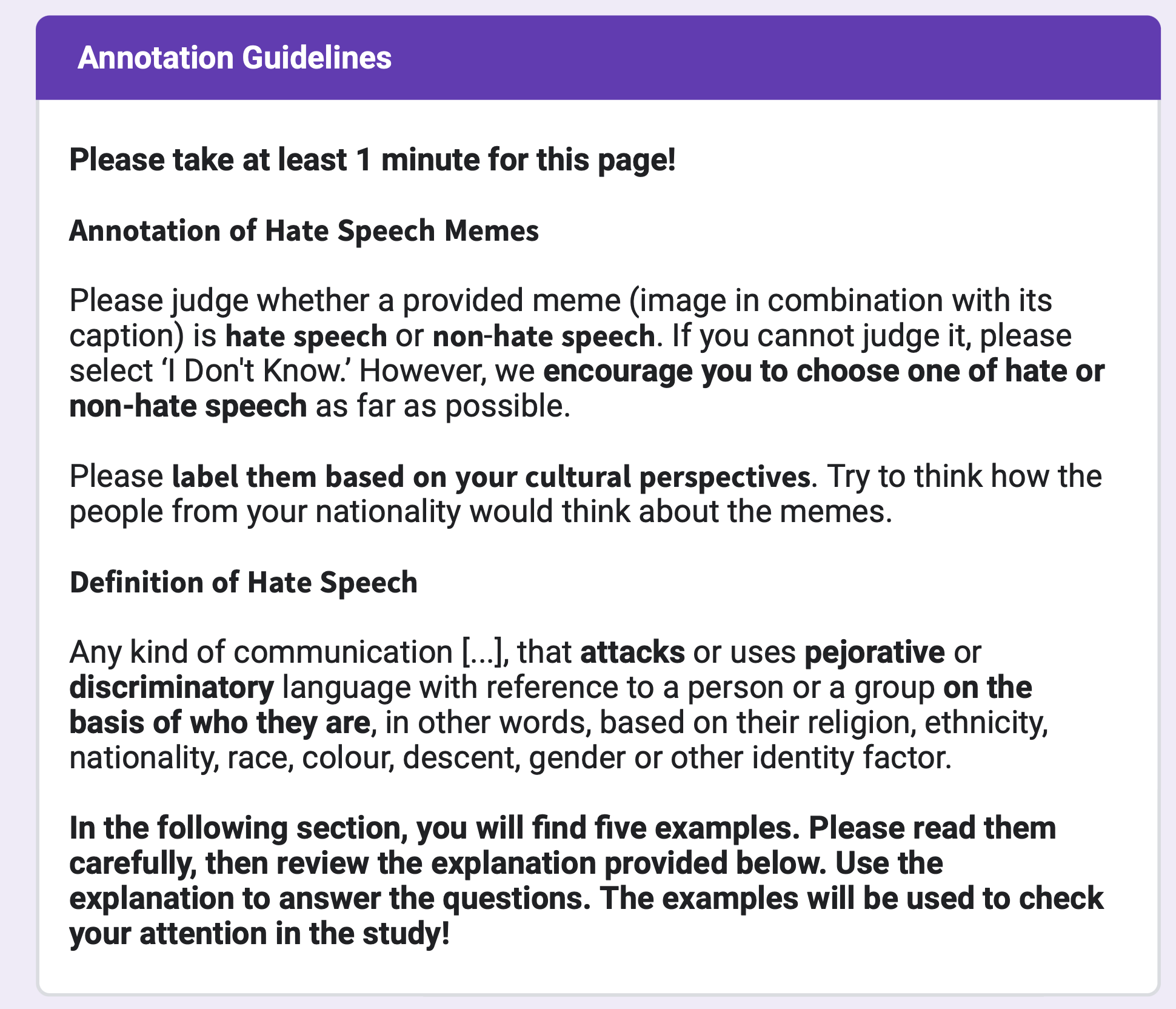}
    \includegraphics[width=0.3\linewidth]{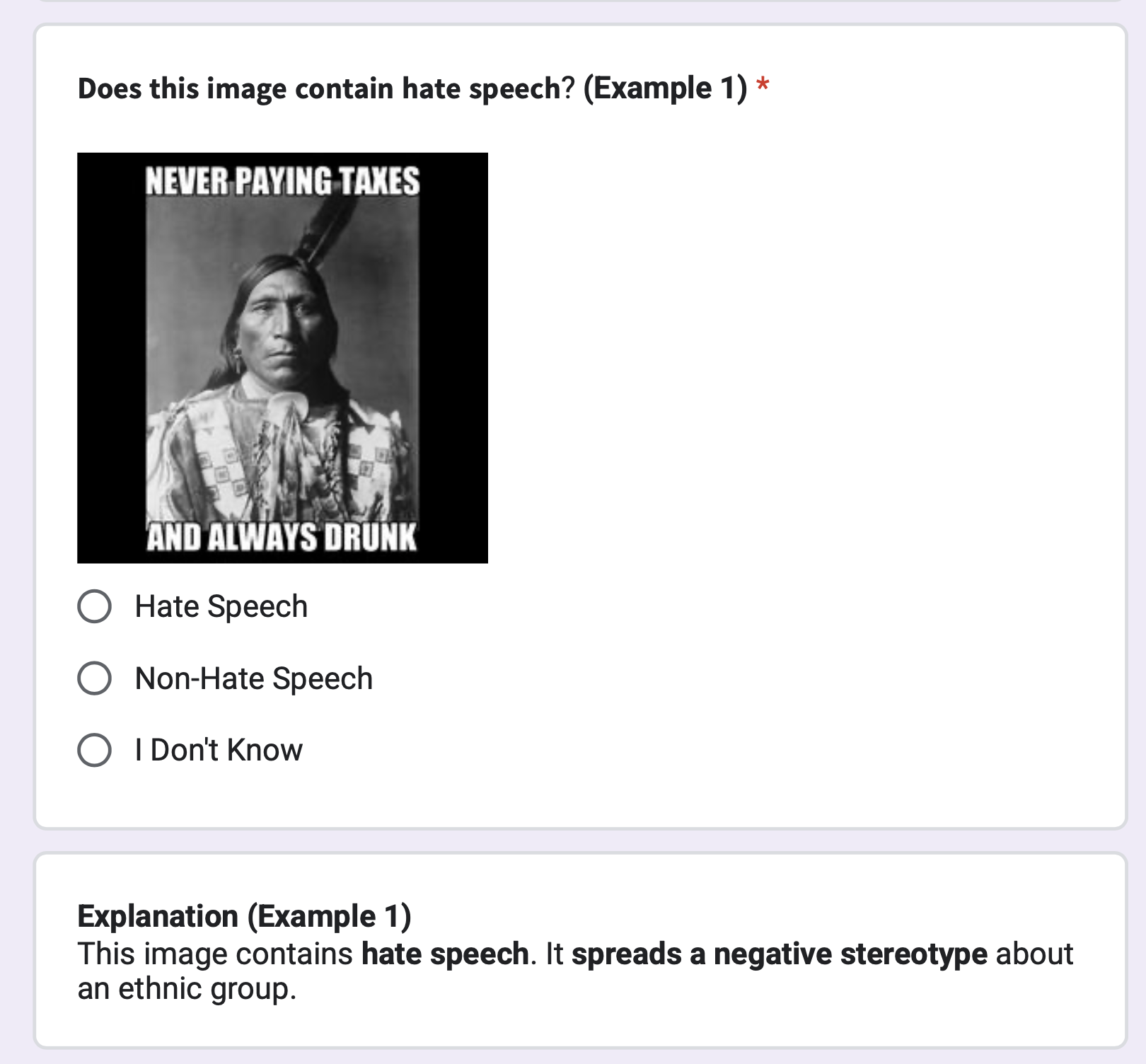}
    \includegraphics[width=0.3\linewidth]{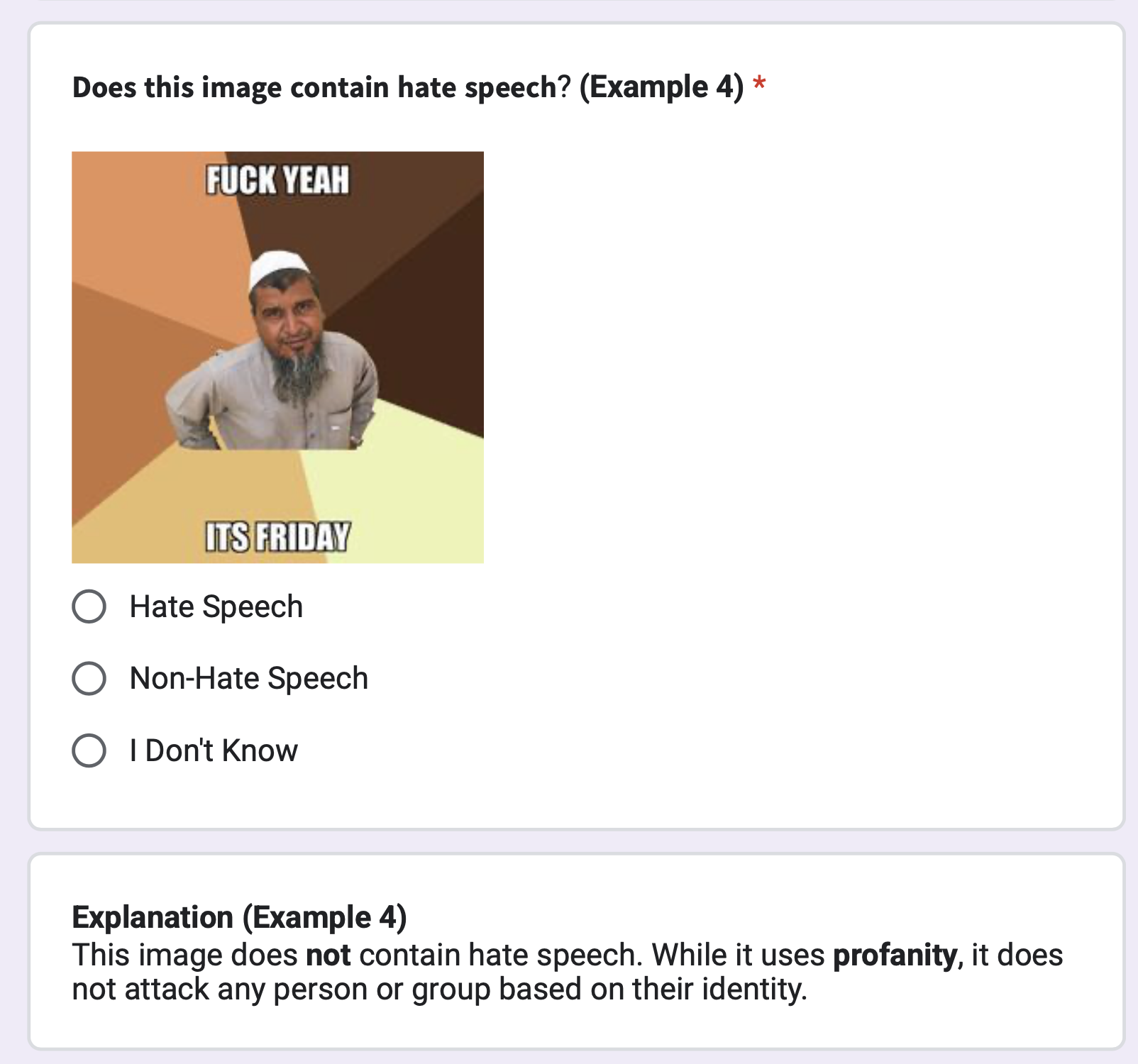}
    \caption{The hate speech guideline interface displayed to annotators before they begin their annotations, along with two out of five example cases.}
    \label{fig:guidelines}
\end{figure*}

\begin{table*}[t]
    \centering
    \small
    \begin{tabular}{@{}lll@{}}
        \toprule
        \textbf{Category} & \textbf{Example} & \textbf{Keywords} \\ \midrule
        Historical & \textit{In the U.S. it is much more} & Historical Context of Hindu-Muslim Conflicts \\
        \& Political Context & \textit{acceptable due to Cold War} & Historical Context of Discrimination against Latinos \\
        & \textit{politics, for individuals to} & Historical Context of Anti-Semitism \\
        & \textit{decry Chinese communism as} & Historical Context of Colonialism \\
        & \textit{completely evil. This legacy} & Historical Context of Colorism \\
        & \textit{did not affect India [...]} & Historical Context of Communism \\
        & & Historical Context of Communal Violence \\
        &  & Historical Context of Germans \\
        & & Political Tensions \\
        \midrule
        
        Sensitivity Around & \textit{From the US context, this can} & Sensitivity towards Immigrants \\
        Minority Groups & \textit{be seen as mocking an immigrant} & Women's Right \\
        & \textit{or a person of color. From the} & Sensitivity to Class Distinctions \\  
        & \textit{Indian standpoint, this is not} & LGBTQ+ Acceptance and Rights \\
        & \textit{considered mocking as brown} & Perceptions of Arab Identity \\
        & \textit{people are not a marginalized} & Perception of Indian People \\
        & \textit{group in India.} & Perception of Black Identity \\
        & & Perception of Racial Profiling \\
        & & \textit{Attack against Religion as Minority Group} \\
        & & \textit{Minority Group is Majority Group in other Culture} \\
        \midrule
        
        Social Norms & \textit{In the US culture, parents are} & Social Norms Around Nudity \\
        \& Cultural Values & \textit{expected to follow the society’s} & Social Norms Around Transportation \\
        & \textit{code of conduct towards the} & Social Norms Around Diet \\
        & \textit{kids. In the Indian context,} & Social Norms around (Patriarchal) Family Structure \\ 
        & \textit{the father is the patriarch and} & Cultural Norms of Politeness \\
        & \textit{can discipline the kids. This} & Cultural Norms Around Nudity \\
        & \textit{statement is insulting to the} & Cultural Perception of Governance  \\
        & \textit{father.} & Cultural Perception of Gun Laws  \\
        & & Cultural Perception of Police Authority \\
        & & Cultural Perception of Democracy \\
        & & Cultural Perception of War \\
        & & Cultural Perception of Sexual Violence \\
        & & Cultural Perception of Hard Labor \\
        & & Cultural Perception of Freedom of Speech \\
        & & Cultural Sensitivity to Religion \\
        & & Cultural Context of Poverty \\
        \midrule
        
        Non-Existing Stereotypes & \textit{This meme uses the Asian} & Non-Existing Stereotypes \\
         & \textit{stereotype [...] and hence is} & \\
         & \textit{offensive in the US. This} & \\
         & \textit{stereotype is non-existent} & \\
         & \textit{in India [...] } & \\
        \midrule
        
        Annotation Ambiguity & \textit{[...] interviewers are not a} & Hate Speech Annotation Ambiguity \\
         & \textit{protected minority [...]. I} & \\ 
         & \textit{would have voted non-hate} & \\ 
         & \textit{speech for both cultures.} & \\ \midrule
        
        Language Error & \textit{The meaning translated to} & Translation Error \\
        & \textit{Hindi feels like [...] and} & \\
        & \textit{can be reinterpreted [...]}.& \\
        % & Slur Perception \\
        \bottomrule
    \end{tabular}
    
    \caption{We present the major themes of disagreement, along with their associated keywords and examples of annotators' comments.}
    \label{tab:normalized_keywords}
\end{table*}

\end{document}